\documentclass{article}

\PassOptionsToPackage{numbers,sort&compress}{natbib}
\usepackage[eandd, final]{neurips_2026}

\usepackage[utf8]{inputenc}
\usepackage[T1]{fontenc}
\usepackage{hyperref}
\usepackage{url}
\usepackage{booktabs}
\usepackage{amsfonts}
\usepackage{nicefrac}
\usepackage{microtype}
\usepackage[table]{xcolor}

\usepackage{graphicx}
\usepackage{subcaption}
\usepackage{makecell}
\usepackage{pgfplots}
\pgfplotsset{compat=1.18}
\usepackage{float}
\usepackage{wrapfig}
\usepackage{enumitem}
\usepackage[most]{tcolorbox}
\usepackage{amsmath}
\usepackage{amssymb}
\usepackage{mathtools}
\usepackage{amsthm}
\usepackage[capitalize,noabbrev]{cleveref}

\newtcolorbox{promptbox}{
  colback=gray!5, colframe=gray!50, boxrule=0.5pt,
  fontupper=\small, left=4pt, right=4pt, top=4pt, bottom=4pt
}

\newtcolorbox{snapbox}[2][]{%
  colback=white, colframe=gray!55, boxrule=0.4pt, arc=2pt,
  fonttitle=\bfseries\small, title=#2, coltitle=black, colbacktitle=gray!12,
  width=\linewidth, top=4pt, bottom=4pt, left=5pt, right=5pt,
  lefttitle=4pt, righttitle=4pt,
  before skip=2pt, after skip=2pt,
  #1
}

\newtcolorbox{episodebox}{%
  colback=white, colframe=gray!55, boxrule=0.4pt, arc=2pt,
  width=\linewidth, top=4pt, bottom=4pt, left=5pt, right=5pt,
  before skip=2pt, after skip=2pt,
}

\usepackage{soul}
\soulregister{\emph}{1}                 % let \hl handle italic emph cleanly
\soulregister{\textit}{1}
\soulregister{\texttt}{1}               % let \hl handle teletype cleanly
\definecolor{precolor}{HTML}{cfe2f3}   % pre-change value (blue)
\definecolor{rulecolor}{HTML}{d9ead3}  % conditional rule (green)
\definecolor{chgcolor}{HTML}{d9d2e9}   % change event (purple)
\definecolor{goldcolor}{HTML}{fce5cd}  % gold answer (beige)
\colorlet{tintraw}{green!15}
\colorlet{tintllm}{blue!15}
\colorlet{tintfile}{orange!15}
\newcommand{\hPre}[1]{{\sethlcolor{precolor}\hl{#1}}}
\newcommand{\hRule}[1]{{\sethlcolor{rulecolor}\hl{#1}}}
\newcommand{\hChg}[1]{{\sethlcolor{chgcolor}\hl{#1}}}
\newcommand{\hGold}[1]{{\sethlcolor{goldcolor}\hl{#1}}}
\newcommand{\stagepass}{\textcolor{green!55!black}{$\checkmark$}}
\newcommand{\stagefail}{\textcolor{red!75!black}{$\boldsymbol{\times}$}}
\newcommand{\pmsd}[1]{{\scriptsize$\pm$#1}}

\theoremstyle{plain}

\theoremstyle{definition}

\theoremstyle{remark}

\title{MEME: Multi-entity \& Evolving Memory Evaluation}

\author{%
  \mdseries
  \textbf{Seokwon Jung}$^{1}$\hspace{1.5em}
  \textbf{Alexander Rubinstein}$^{2,*}$\hspace{1.5em}
  \textbf{Arnas Uselis}$^{2,*}$\\[2pt]
  \textbf{Sangdoo Yun}$^{3}$\hspace{1.5em}
  \textbf{Seong Joon Oh}$^{1}$\\[6pt]
  $^{1}$KAIST \hspace{1.5em}
  $^{2}$T\"ubingen AI Center, University of T\"ubingen \hspace{1.5em}
  $^{3}$NAVER AI Lab\\[2pt]
  \texttt{SeokwonJung@kaist.ac.kr}
}

\begin{document}

\maketitle

{\renewcommand\thefootnote{*}\footnotetext{Equal contribution.}}

\begin{abstract}
LLM-based agents increasingly operate in persistent environments where they must store, update, and reason over information across many sessions. While prior benchmarks evaluate only single-entity updates, MEME defines six tasks spanning the full space defined by the multi-entity and evolving axes, including three not scored by prior work: Cascade and Absence (dependency reasoning) and Deletion (post-removal state). Evaluating six memory systems spanning three memory paradigms on 100 controlled episodes, we find that all systems collapse on dependency reasoning under the default configuration (Cascade: 3\%, Absence: 2\% in average accuracy) despite adequate static retrieval performance. Prompt optimization, deeper retrieval, reduced filler noise, and most stronger LLMs fail to close this gap. The largest gain comes from a file-based agent paired with Claude Opus 4.7 as its internal LLM, which partially closes the gap at roughly 23$\times$ the baseline cost. Code and data are available on the project page: \url{https://seokwonjung-jay.github.io/meme-eval/}.
\end{abstract}

\section{Introduction}

\setlength{\intextsep}{6pt}
\setlength{\columnsep}{12pt}
\begin{wrapfigure}[18]{r}{0.42\linewidth}
\vspace{-10pt}
\centering
    \includegraphics[width=\linewidth]{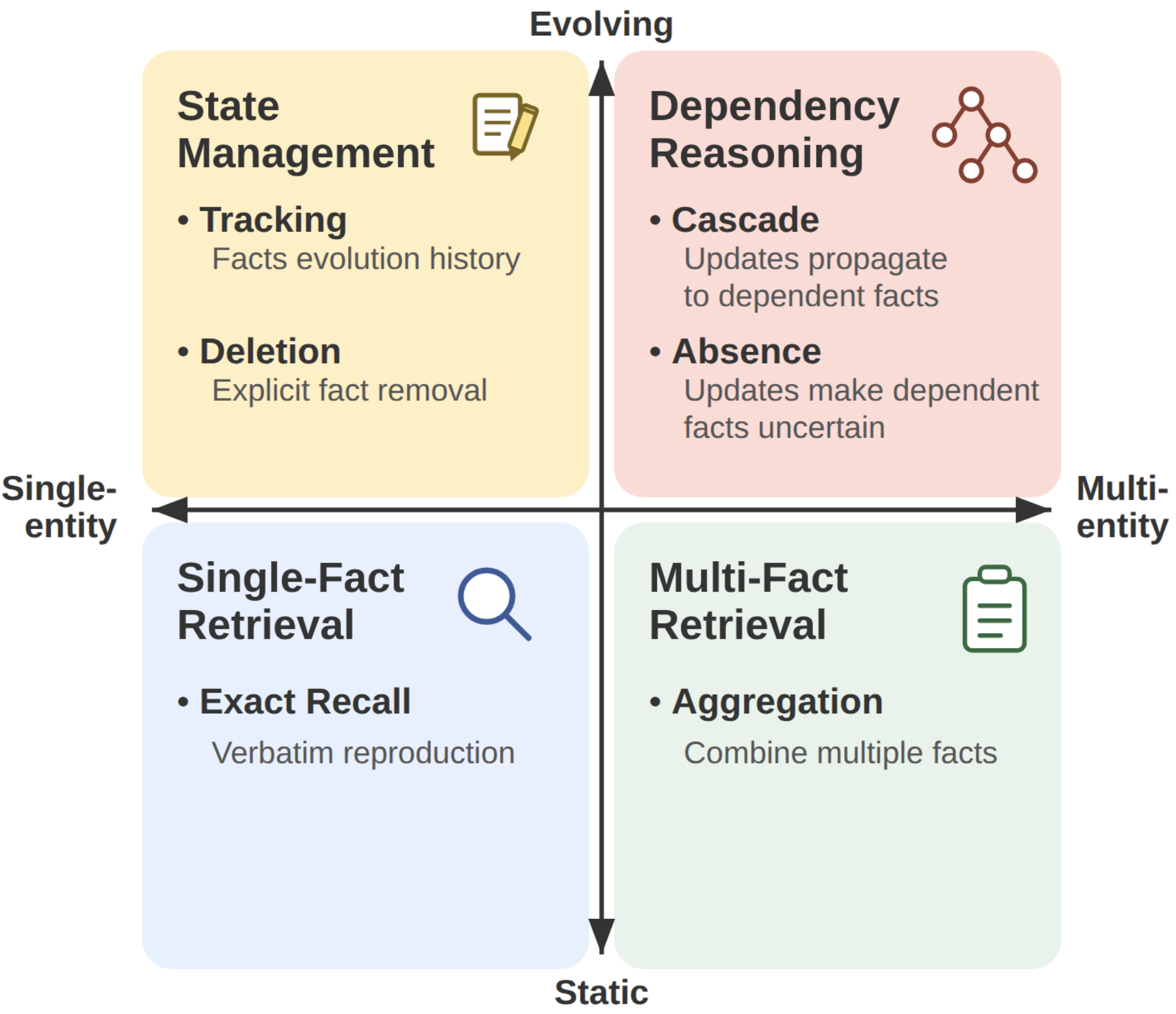}
\caption{MEME's taxonomy of memory operations along two dimensions: entity scope (single vs.\ multi-entity) and temporal dynamics (static vs.\ evolving), with six tasks distributed across the four quadrants.}
\label{fig:Tasktaxonomy}
\end{wrapfigure}

As large language models (LLMs) increasingly serve as agents that interact with users across many sessions, accurately storing, updating, and reasoning over past interactions has become essential \cite{zhang2024survey}. For instance, when a user reports moving to a new city, the agent must not only record this change but also recognize that previously stored facts that depended on the old residence, such as commute time or nearby facilities, may no longer be valid. Today's memory systems address such needs through three broad paradigms: raw retrieval over unprocessed text chunks \cite{izacard2022contriever}, LLM-processed memory that extracts and reorganizes facts \cite{chhikara2025mem0, packer2023memgpt, rasmussen2025zep, edge2024graphrag}, and file-based agents where an LLM manages persistent files via tool-calling \cite{karpathy2025wiki}. Evaluation for these systems has evolved from single-turn long-context benchmarks \cite{hsieh2024ruler, modarressi2025nolima} through multi-session evaluations of static fact retention \cite{maharana2024locomo} to benchmarks with dynamic updates, where entity values change across sessions \cite{wu2024longmemeval, tan2025membench, hu2025memoryagentbench}.

Yet existing memory benchmarks evaluate updates only for independent entities, missing the dependency reasoning that real interactions require (\cref{tab:benchmark_comparison}). No prior benchmark scores how a dependent fact changes after an upstream update (Cascade), how a previously valid answer becomes uncertain (Absence), or how a removed fact stops being reported (Deletion). This leaves a critical blind spot in how today's memory systems handle stateful, interdependent knowledge.

\begin{wraptable}{r}{0.55\linewidth}
\centering
\caption{Comparison between MEME and existing memory benchmarks. Tasks: Exact Recall (ER), Aggregation (Agg), Tracking (Tr), Deletion (Del), Cascade (Cas), Absence (Abs). Mappings are approximated from reported task types. \textcolor{green!60!black}{\checkmark}~coverage; \textcolor{red}{$\times$}~absence.}
\label{tab:benchmark_comparison}
\footnotesize
\renewcommand{\arraystretch}{0.85}
\setlength{\tabcolsep}{3pt}
\begin{tabular*}{\linewidth}{@{\extracolsep{\fill}}l*{6}{c}@{}}
\toprule
& \textbf{ER} & \textbf{Agg} & \textbf{Tr} & \textbf{Del} & \textbf{Cas} & \textbf{Abs} \\
\midrule
RULER \cite{hsieh2024ruler}              & \textcolor{green!60!black}{\checkmark} & \textcolor{green!60!black}{\checkmark} & \textcolor{red}{$\times$} & \textcolor{red}{$\times$} & \textcolor{red}{$\times$} & \textcolor{red}{$\times$} \\
NoLiMa \cite{modarressi2025nolima}       & \textcolor{green!60!black}{\checkmark} & \textcolor{red}{$\times$} & \textcolor{red}{$\times$} & \textcolor{red}{$\times$} & \textcolor{red}{$\times$} & \textcolor{red}{$\times$} \\
LoCoMo \cite{maharana2024locomo}         & \textcolor{green!60!black}{\checkmark} & \textcolor{green!60!black}{\checkmark} & \textcolor{red}{$\times$} & \textcolor{red}{$\times$} & \textcolor{red}{$\times$} & \textcolor{red}{$\times$} \\
LongMemEval \cite{wu2024longmemeval}     & \textcolor{green!60!black}{\checkmark} & \textcolor{green!60!black}{\checkmark} & \textcolor{green!60!black}{\checkmark} & \textcolor{red}{$\times$} & \textcolor{red}{$\times$} & \textcolor{red}{$\times$} \\
MemBench \cite{tan2025membench}          & \textcolor{green!60!black}{\checkmark} & \textcolor{green!60!black}{\checkmark} & \textcolor{green!60!black}{\checkmark} & \textcolor{red}{$\times$} & \textcolor{red}{$\times$} & \textcolor{red}{$\times$} \\
MemoryAgentBench \cite{hu2025memoryagentbench} & \textcolor{green!60!black}{\checkmark} & \textcolor{red}{$\times$} & \textcolor{green!60!black}{\checkmark} & \textcolor{red}{$\times$} & \textcolor{red}{$\times$} & \textcolor{red}{$\times$} \\
\midrule
\textbf{MEME (Ours)} & \textcolor{green!60!black}{\checkmark} & \textcolor{green!60!black}{\checkmark} & \textcolor{green!60!black}{\checkmark} & \textcolor{green!60!black}{\checkmark} & \textcolor{green!60!black}{\checkmark} & \textcolor{green!60!black}{\checkmark} \\
\bottomrule
\end{tabular*}
\end{wraptable}

\begin{figure*}[!t]
\centering
    \includegraphics[width=1\linewidth]{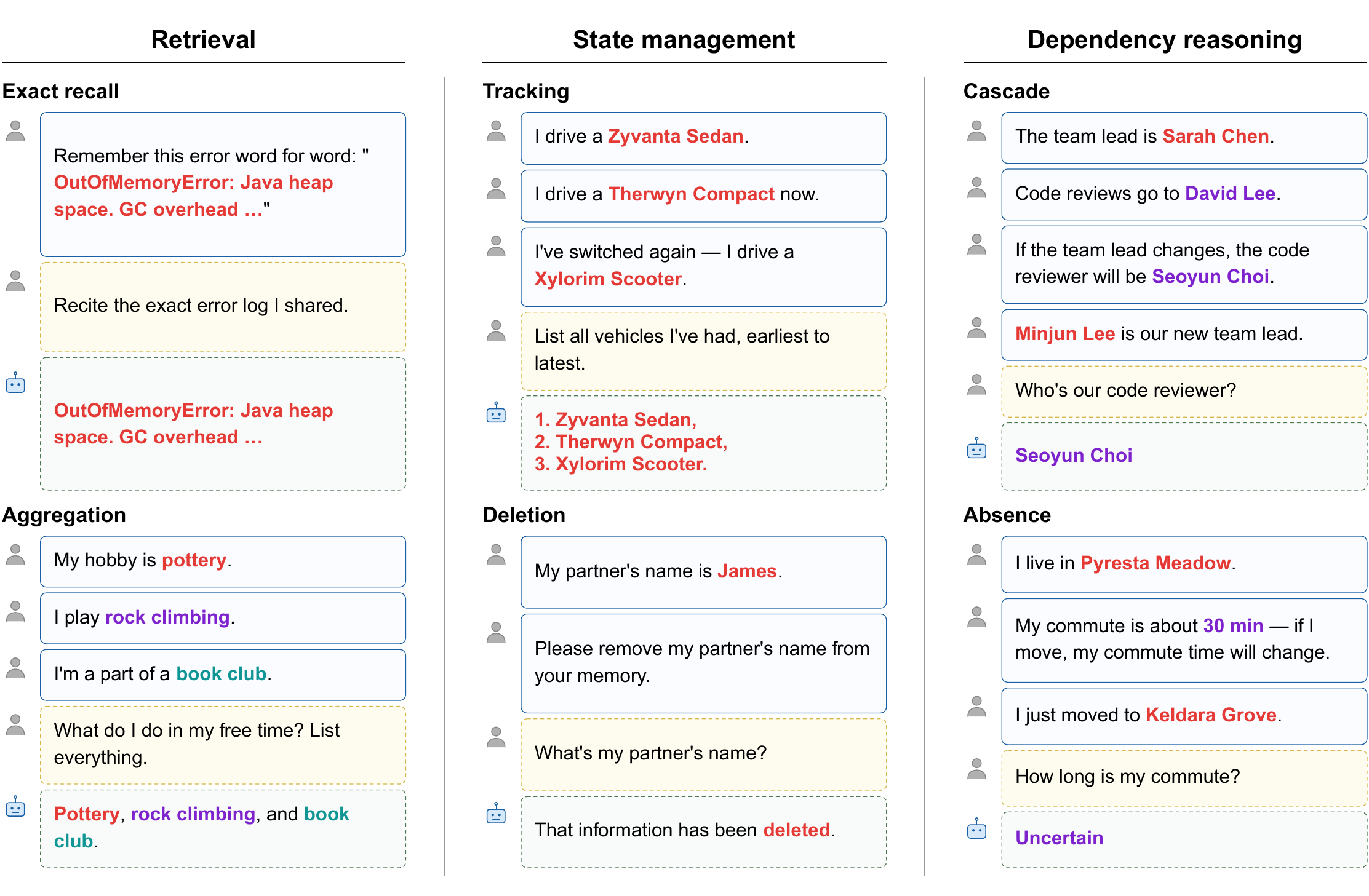}

\caption{Examples of the six MEME task types across three categories: \textbf{Left:} Retrieval (Exact Recall, Aggregation), which merges the Single-Fact Retrieval and Multi-Fact Retrieval quadrants of \cref{fig:Tasktaxonomy}. \textbf{Middle:} State Management (Tracking, Deletion). \textbf{Right:} Dependency Reasoning (Cascade, Absence). Each distinct entity is rendered in a different color.}
\label{fig:task_types}
\end{figure*}

To address this gap, we argue that a complete memory evaluation must be organized along two orthogonal dimensions: \emph{entity scope} (single vs.\ multi-entity) and \emph{temporal dynamics} (static vs.\ evolving). These dimensions reflect known challenges in related fields: the entity scope axis parallels the single-hop vs.\ multi-hop distinction in question answering \cite{yang2018hotpotqa, trivedi2022musique}, while the temporal dynamics axis parallels the ripple effect problem in knowledge editing, where modifying one fact requires propagating changes to logically dependent facts \cite{zhong2023mquake, cohen2024ripple}. While these axes have been studied separately, real interactions combine them: an update to one entity can ripple through multiple dependents over time, making joint evaluation along both axes essential.

Based on this framework, we present MEME (\textbf{M}ulti-entity and \textbf{E}volving \textbf{M}emory \textbf{E}valuation), a benchmark that defines six tasks targeting memory-intensive operations in each quadrant of this two-dimensional space (\cref{fig:Tasktaxonomy}). Our contributions are:

\begin{itemize}
  \item \textbf{A principled evaluation taxonomy.} We organize memory evaluation along entity scope $\times$ temporal dynamics and select representative tasks per quadrant, including \emph{Cascade} (inferring unstated changes from dependency rules), \emph{Absence} (recognizing that a previously valid answer is no longer trustworthy), and \emph{Deletion} (verifying that a removed fact is no longer reported), task types that no existing benchmark scores.
  \item \textbf{A rigorously controlled dataset with verifiable and solvable ground truth.} We generate episodes from a DAG-based knowledge graph with explicit conditional rules across two domains (Personal Life and Software Project); the DAG structure makes gold answers verifiable by construction, and an in-context validation (gold facts fed directly to the answering LLM) confirms the tasks are solvable in principle.
  \item {\textbf{A diagnostic study of where current memory systems fail and where partial closure emerges.} Evaluating six systems across three architectural paradigms, we find: (i) under default configurations every system fails Cas/Abs (Cascade: 0.03, Absence: 0.02 in average accuracy), a gap that persists under prompt optimization, deeper retrieval, a stronger answering LLM, or reduced filler noise; (ii) most systems store both the dependency rule and the change event, yet fail to surface the change at retrieval or, when it is retrieved, to use it when answering; (iii) partial closure emerges when MD-flat uses Opus 4.7 as its internal LLM, where this internal LLM writes the propagated value into the store at ingest so the retriever surfaces it directly, but at ${\sim}23\times$ the baseline cost.}
\end{itemize}

\section{Related Work}
\label{sec:related}

We review prior work on memory system architectures and on the benchmarks used to evaluate them, framing the gap that MEME addresses.

{
\textbf{Memory architectures.} LLM memory systems span three paradigms. Raw retrieval stores session text as chunks and retrieves via lexical (BM25) or semantic similarity \cite{izacard2022contriever}, preserving original utterances but bounded by a fixed top-$k$ window. LLM-processed memory uses an internal LLM during ingestion to extract or restructure: Mem0 \cite{chhikara2025mem0} decomposes conversations into atomic facts \cite{lewis2020rag}, MemGPT \cite{packer2023memgpt} pages between working memory and external storage, Graphiti \cite{rasmussen2025zep}, the open-source temporal-knowledge-graph engine underlying Zep, encodes entity-relation triples, and GraphRAG \cite{edge2024graphrag} adds community summaries. File-based agents hand LLMs tool-calling access to persistent markdown stores curated across sessions, including Hermes\footnote{\url{https://github.com/NousResearch/hermes-agent}}, OpenClaw\footnote{\url{https://github.com/openclaw/openclaw}}, and the Karpathy Wiki \cite{karpathy2025wiki}. These three paradigms span the systems we evaluate; we benchmark all of them on dependency reasoning and find that none closes the gap under default configurations.

\textbf{Memory benchmarks.} Stateless probes like RULER \cite{hsieh2024ruler} and NoLiMa \cite{modarressi2025nolima} measure attention-window limits within a single input rather than persistent memory across sessions. Multi-session benchmarks evaluate memory across sessions: LoCoMo \cite{maharana2024locomo} tests retention of static preferences, while LongMemEval \cite{wu2024longmemeval}, MemBench \cite{tan2025membench}, and MemoryAgentBench \cite{hu2025memoryagentbench} extend evaluation to evolving memory through knowledge updates, abstention, aggregation, and selective forgetting. These tasks remain isolated, single-entity updates and do not evaluate the ripple effects an upstream change should trigger in dependent entities. MEME differs by scoring three task types absent in prior work (Cascade, Absence, Deletion), which are constructed from a DAG-based knowledge graph with verifiable propagation gold answers.
}

\section{MEME}

MEME maps the entity-scope $\times$ temporal-dynamics framework to six tasks (\cref{sec:task-defs}) and a DAG-based generation pipeline that yields verifiable gold answers across 100 episodes (\cref{sec:dataset-gen}).

\subsection{Task Definitions}\label{sec:task-defs}

Within each quadrant of the entity scope $\times$ temporal dynamics space (\cref{fig:Tasktaxonomy}), we select one or two memory-intensive operations from those commonly encountered in long-running agent deployments as the representative task(s); easier variants are already covered by existing benchmarks and are intentionally excluded. Concrete examples of all six tasks are illustrated in \cref{fig:task_types}.

\textbf{Exact Recall (ER)} targets a single static entity and demands character-level verbatim reproduction, testing encoding fidelity. \textbf{Aggregation (Agg)} combines multiple static entities scattered across separate sessions into a single answer, testing retrieval coverage when no explicit link connects them. \textbf{Tracking (Tr)} reconstructs the full revision history of a single evolving entity in chronological order, testing whether past values are retained rather than overwritten. \textbf{Deletion (Del)} tests whether the system stops reporting a fact after the user explicitly removes it, rather than continuing to surface the old value. \textbf{Cascade (Cas)} infers that a dependent entity's value has changed based on a stated dependency rule and an upstream update, testing propagation through dependency chains. \textbf{Absence (Abs)} recognizes that a dependent entity is uncertain after an upstream change with no replacement rule, where the correct answer is uncertainty rather than a new value.

\subsection{Dataset Generation}\label{sec:dataset-gen}

We generate the MEME dataset across two domains, Personal Life (PL; everyday interactions with a personal assistant) and Software Project (SW; collaborative planning of a software project), in two stages. First, we define a knowledge graph per domain that encodes entities and their dependencies. Then, we construct evaluation episodes by sampling entity subsets, assigning tasks, verbalizing facts into conversations, and assembling filler-interleaved haystacks.

\paragraph{Knowledge graph.}

Each domain is built on a Directed Acyclic Graph (DAG) $G = (V, E, \mathcal{P}, \Phi)$, where $V$ is a set of entities (e.g., \emph{health\_condition}, \emph{medication}), $E \subseteq V \times V$ contains directed dependency edges (e.g., \emph{health\_condition} $\to$ \emph{medication}), $\mathcal{P}(v)$ is the value pool for entity $v$, and $\Phi$ is a set of conditional rules. Each rule $\phi_v \in \Phi$ specifies how a descendant $v$'s value depends on its parents' (e.g., ``if health condition changes to high blood pressure, switch medication to Thrynexol'').

The dataset comprises 100 evaluation episodes (50 per domain). Each domain uses a single hand-crafted knowledge graph $G$ reused across episodes (Personal Life: 39 entities, 34 edges; Software Project: 51, 27; full breakdown in \cref{tab:dataset_stats_detail}). Each episode is $\sim$35K tokens of dialogue context, and the dataset yields 684 post-change evaluation questions across the six task types (328 PL + 356 SW).\footnote{This count excludes ten Absence questions that are mislabeled in the v1.0 data release: in these episodes the sessions also state the replacement value, so the gold answer should be that value rather than ``Uncertain''. All results in this version exclude them; earlier versions of this paper included them.} All entity values use fictitious names to prevent parametric knowledge contamination; graph details in \cref{app:dataset}.

\paragraph{Episode construction.}

Each episode is a tuple $\epsilon = (G, \mathcal{S}, \mathcal{Q}, \mathcal{A})$, where $\mathcal{S}$ is a chronological sequence of conversational sessions, $\mathcal{Q}$ is the set of evaluation questions, and $\mathcal{A}$ is the corresponding gold answers. For Cascade and Absence tasks, the gold answer is not stated in $\mathcal{S}$ but is computed by propagating updates through $G$. When a parent $v_i$ is updated in $\mathcal{S}$, the resolved state of dependent $v$ is:
\begin{equation}
v^* = \begin{cases} \phi_v(v_i^*) & \text{if } \phi_v \in \Phi \quad \text{(Cascade)} \\ \bot & \text{if } \phi_v \notin \Phi \quad \text{(Absence)} \end{cases}
\end{equation}
Here $\bot$ denotes that no answer is derivable from the available rules; the gold answer for Absence is ``Uncertain''. The definition is recursive: for a chain $r \to v_1 \to v_2$, $v_1^* = \phi_{v_1}(r^*)$ and $v_2^* = \phi_{v_2}(v_1^*)$, so a single root change propagates through multi-hop chains. The gold answer for these tasks is $a = v^*$. We refer to the value of $v$ stated in $\mathcal{S}$ before the upstream change as the \emph{pre-change value}, in contrast to the resolved $v^*$.

We construct each episode in five steps over the fixed graph $G$:
\begin{enumerate}[leftmargin=1.6em,itemsep=2pt,topsep=2pt]
  \item \textit{Entity set selection.} A root entity is selected from $G$. The episode uses this root, its descendants in $G$, and a sample of entities from outside its cascade chain.
  \item \textit{Value assignment.} Each entity in the episode is assigned an initial value from its value pool $\mathcal{P}(v)$. Domain-specific consistency constraints are applied as a post-processing pass to ensure the initial graph state is logically coherent. For example, if \emph{vehicle} is \emph{none}, \emph{commute\_method} excludes \emph{driving}.
  \item \textit{Task assignment.} Entities are mapped to task types based on their topological role in $G$:
    \begin{itemize}[leftmargin=1.4em,itemsep=1pt,topsep=1pt]
      \item Tracking: entities outside the cascade chain, with three value updates across the episode.
      \item Cascade and Absence: sampled from the root's descendants.
      \item Aggregation: predefined triples drawn from descendants and entities outside the cascade chain.
      \item Exact Recall and Deletion: entities outside the cascade chain.
    \end{itemize}
  \item \textit{Verbalization.} We employ a hybrid approach to convert the structured skeleton into conversational sessions. Base facts are converted into multi-turn dialogues via LLM self-chat (gpt-4o), where a User LLM and an Assistant LLM alternate turns to produce natural conversation from structured fact seeds (full session in \cref{app:example-session}). In contrast, dependency rules and exact recall facts are embedded using template-direct (verbatim) text to ensure absolute factual precision. A two-layer LLM verification pass (gpt-4o annotation, Gemini~2.5 Flash semantic audit) confirms that all self-chat turns faithfully reflect the underlying gold facts (details in \cref{app:prompts-verify}).
  \item \textit{Haystack assembly.} Evidence sessions $S_{\text{ev}}$ are interleaved with filler sessions $S_{\text{fill}}$. To prevent semantic interference, we use an offline pre-processing pipeline where a domain-matched corpus is filtered using a hybrid retrieval-and-conflict-removal strategy (BM25 and text-embedding-3-small surface candidates for a gpt-4o-mini conflict judge). During final assembly, we apply a keyword-based blocklist to select pre-filtered fillers that do not clash with the gold facts of the current episode. The resulting episodes contain approximately 35,000 tokens. Full filtering pipeline, statistics, and rejection examples are in \cref{app:haystack}.
\end{enumerate}

\section{Experiments}

We evaluate six memory systems on MEME and find that all of them fail Cascade and Absence. We then ask two questions in turn: \emph{where} in each system the dependency information is lost (\cref{sec:where-loss}), and \emph{whether} we can close the gap without changing the memory architecture (\cref{sec:no-closure}). One configuration partially closes the gap; we examine what made it possible and what it costs.

\subsection{Experimental Setup}
\label{sec:setup}

\paragraph{Systems and LLM roles.}
We evaluate six memory systems spanning the three paradigms identified in \cref{sec:related}: \emph{raw retrieval} (BM25 \cite{lu2024bm25s}, text-embedding-3-small), \emph{LLM-processed memory} (Mem0 \cite{chhikara2025mem0}, Graphiti \cite{rasmussen2025zep}), and \emph{file-based agents} (Karpathy Wiki \cite{karpathy2025wiki} and MD-flat). Per-system configurations are in \cref{app:systems}; ingestion, retrieval, and answer prompts are in \cref{app:prompts}. The Karpathy Wiki uses an LLM to extract knowledge from each session into dated daily logs and periodically compiles those logs into topic-specific concept articles that retrieval reads from, while MD-flat (our minimal single-file baseline) keeps all facts in a single markdown file curated through read/write/append tool calls. All systems ingest identical chronological session transcripts and use gpt-4.1-mini uniformly in two roles: as the \emph{internal LLM} (used inside the memory system for ingestion, extraction, or retrieval planning) and as the \emph{answering LLM} (which produces the final user-facing answer from retrieved context). This places every system on the same language-model footing and isolates differences in memory architecture rather than in answer-time reasoning. Five systems issue the two roles as separate LLM calls and share one single-step answer prompt that asks for the value only (\cref{app:prompts}), while Karpathy Wiki performs both within a single multi-step agentic loop; \cref{sec:no-closure} tests a task-specific answer prompt separately. We additionally include an \emph{in-context} baseline that bypasses the memory system and feeds the entire 32K-filler episode transcript directly to the answering LLM (gpt-4.1-mini and Sonnet 4.6). This baseline anchors the cost-efficiency reference for memory architectures, which trade one-time ingestion overhead for cheaper per-query inference.

{\paragraph{Memory pipeline.}\label{sec:setup-frame}
We refer to three stages within each memory system that we will reuse throughout the analysis: \emph{encoding} (writing each user-stated fact and conditional rule into the store at ingestion), \emph{maintenance} (retaining the rule and any subsequent change events in the store up to query time), and \emph{retrieval} (surfacing that content for the answering LLM at query time). \Cref{tab:pipeline-overview} maps each system across these stages plus its storage substrate.}

\paragraph{Evaluation protocol.}
Answer correctness is evaluated by a GPT-4o judge \cite{zheng2023llmjudge}, validated against annotations by one of the authors on 144 samples (98.6\% agreement, Cohen's $\kappa$ \cite{cohen1960kappa} of 0.965); task-specific judge prompts are in \cref{app:prompts-judge}. For Cascade, Absence, and Deletion tasks, we apply \emph{trivial-pass filtering}: credit requires correct answers both before and after the change or delete event. For example, on a Deletion task where the user first says their hobby is pottery and later asks to remove that fact, the system is credited only if it recalls pottery beforehand and stops reporting it afterward. This excludes false positives from systems that never encoded the fact. A gold-facts in-context ceiling, where only task-relevant gold facts are fed directly to the answering LLM, confirms that the tasks are solvable in principle: 0.91 overall with Claude Opus 4.7 (full breakdown across four answer LLMs in \cref{app:in-context-ceiling}). We compute per-episode dollar cost from observed LLM token usage at each LLM's public per-token rate, reported separately for ingestion and for inference (the retrieval and answer stages); \cref{app:costs} provides the per-stage breakdown. Overall is the unweighted mean of the six task accuracies.

\subsection{Main Results}
\label{sec:main-results}

\begin{table*}[t]
\centering
\caption{System accuracy across the six MEME task types, grouped by paradigm. All memory systems run end-to-end on gpt-4.1-mini (both memory operations and answer generation). Best per task in \textbf{bold}. Cascade, Absence, and Deletion use trivial-pass filtering. The \emph{Average} row reports the mean over the six memory systems (the in-context paradigm is excluded).}
\label{tab:main_results}
\footnotesize
\setlength{\tabcolsep}{4pt}
\renewcommand{\arraystretch}{0.85}
\begin{tabular*}{\textwidth}{@{}@{\extracolsep{\fill}}l*{7}{c}*{2}{c}@{}}
\toprule
& \multicolumn{7}{c}{\textbf{Accuracy}} & \multicolumn{2}{c}{\textbf{Cost \$/ep}} \\
\cline{2-8}\cline{9-10}
\addlinespace[3pt]
& \textbf{ER} & \textbf{Agg} & \textbf{Tr} & \textbf{Del} & \textbf{Cas} & \textbf{Abs} & \textbf{Overall} & \textbf{Ingest} & \textbf{Inference} \\
\midrule
\multicolumn{10}{l}{\textit{In-context (no memory)}} \\
~~gpt-4.1-mini  & 1.00 & 0.27 & 0.69 & 0.45 & 0.03 & 0.03 & 0.41 & - & \$0.16 \\
~~Sonnet 4.6    & 0.50 & 0.21 & 0.58 & 0.39 & 0.05 & 0.34 & 0.35 & - & \$1.50 \\
\midrule
\multicolumn{10}{l}{\textit{Raw retrieval}} \\
~~BM25                   & \textbf{1.00}& 0.05  & 0.16 & \textbf{0.27} & 0.02  & 0.00           & 0.25 & - & \$0.04 \\
~~text-embedding-3-small & 0.96          & 0.33 & 0.46 & 0.17          & 0.04  & 0.00           & 0.33 & - & \$0.03 \\
\midrule
\multicolumn{10}{l}{\textit{LLM-processed memory}} \\
~~Mem0                   & 0.67          & 0.35 & 0.43 & 0.21          & 0.03  & 0.00           & 0.28 & \$0.10 & \$0.01 \\
~~Graphiti               & 0.03           & 0.01  & 0.04  & 0.09           & 0.02  & 0.01           & 0.03 & \$0.55 & \$0.00 \\
\midrule
\multicolumn{10}{l}{\textit{File-based agents}} \\
~~MD-flat                & 0.94          & \textbf{0.45} & \textbf{0.77} & 0.25          & \textbf{0.05} & \textbf{0.06} & \textbf{0.42} & \$0.04 & \$0.01 \\
~~Karpathy Wiki          & 0.11          & 0.18 & 0.27 & 0.03           & 0.01  & 0.03           & 0.10 & \$1.17 & \$0.22 \\
\midrule
\addlinespace[3pt]
\textit{Average} & 0.62 & 0.23 & 0.35 & 0.17 & 0.03 & 0.02 & 0.24 & - & - \\
\bottomrule
\end{tabular*}
\end{table*}

\cref{tab:main_results} summarizes accuracy across all six tasks. We highlight three findings below.

\begin{wrapfigure}{r}{0.5\linewidth}
\vspace{8pt}
\centering
\begin{subfigure}[t]{0.33\linewidth}
\includegraphics[width=\linewidth]{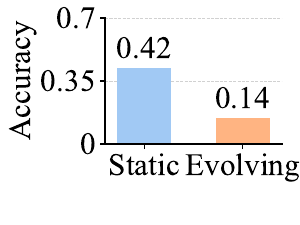}
\caption{temporal axis}
\end{subfigure}\hfill
\begin{subfigure}[t]{0.33\linewidth}
\includegraphics[width=\linewidth]{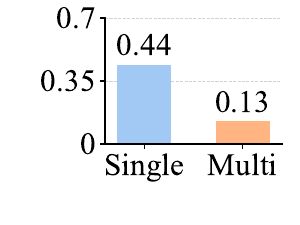}
\caption{entity-scope axis}
\end{subfigure}\hfill
\begin{subfigure}[t]{0.33\linewidth}
\includegraphics[width=\linewidth]{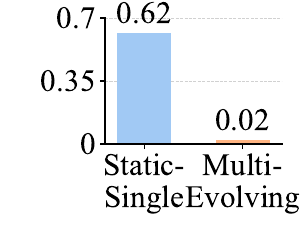}
\caption{both crossed}
\end{subfigure}
\caption{Marginal effect of each evaluation axis on mean accuracy across the six main-table systems. Each axis individually drops accuracy by $\sim$0.30; crossing both leaves the Multi-Evolving cell at a 0.02 floor.}
\label{fig:heatmap-axes}
\end{wrapfigure}

\textbf{No system reliably solves dependency reasoning.} The best system (MD-flat) reaches only 0.42 overall. Dependency reasoning is the most consistent failure: Cascade averages 0.03 and Absence 0.02 across all six systems, well below the per-task averages on every static task (the lowest is Aggregation at 0.23). This failure is consistent across all three paradigms and stable across $N{=}5$ samplings (\cref{app:sd}). It also appears in every slice of the two graphs by domain, hop length, dependency pattern, and root entity, already at single-hop dependencies, with no slice above 0.15 on Cascade or 0.04 on Absence (\cref{app:dag-sensitivity}). Removing the explicit cue from the dependency rules only makes the tasks harder: the memory systems stay near zero, and even the gold-facts ceiling drops sharply on Absence (\cref{app:implicit}).

\textbf{The two evaluation axes shape system performance.} Both axes substantially reduce mean accuracy on their own (entity scope 0.31, temporal 0.28), so neither is redundant; crossing both pushes the Multi-Evolving cell to the floor (0.02, \cref{fig:heatmap-axes}).

\begin{figure}[t]
\centering
{\centering\footnotesize
\colorbox{precolor}{\rule{0pt}{0.7em}\hspace{1em}}~pre-change value\hspace{1em}%
\colorbox{rulecolor}{\rule{0pt}{0.7em}\hspace{1em}}~conditional rule\hspace{1em}%
\colorbox{chgcolor}{\rule{0pt}{0.7em}\hspace{1em}}~change event\hspace{1em}%
\colorbox{goldcolor}{\rule{0pt}{0.7em}\hspace{1em}}~answer\par}\vspace{6pt}

\begin{episodebox}
\footnotesize
\textbf{Episode sw\_033 (Software Project, Cascade).}
\par\setlength{\leftskip}{1em}\setlength{\parskip}{0pt}
Sess.\,11: \emph{``Weekly report recipient is \hPre{Hyunwoo Nam} (assigned by team lead Seokjin Kang). \hRule{If the team lead changes, the recipient will be James Lee.}''}\par
Sess.\,14: \emph{``\hChg{Jihoon Ryu is our new team lead.}''}\par
\vspace{4pt}
Question: \emph{``Who receives the weekly report?''}\quad Answer: \hGold{James Lee}.\par
\end{episodebox}\vspace{3pt}

\begin{minipage}[t]{0.498\textwidth}
\begin{snapbox}[equal height group=fig4eq]{Graphiti (gpt-4.1-mini) -- Failure}
\footnotesize
\textbf{1.\,\stagepass\,Encoding}: all facts encoded\\[2pt]
{\tiny\hangindent=1em\hangafter=1
Seokjin Kang assigns the weekly report recipient as \hPre{Hyunwoo Nam} \ldots\\
\hRule{If the team lead changes, the weekly report recipient will be James Lee \ldots}\par}\vspace{6pt}
\textbf{2.\,\stagepass\,Maintenance}: all facts retained\\[2pt]
{\tiny\hangindent=1em\hangafter=1
\hChg{Jihoon Ryu is the new team lead of the User}\par}\vspace{6pt}
\textbf{3.\,\stagefail\,Retrieval}: change event not surfaced\\[2pt]
{\tiny\hangindent=1em\hangafter=1
\hRule{If the team lead changes, \ldots James Lee \ldots}\\
\hPre{Hyunwoo Nam} \ldots\\
\mbox{[unrelated filler]}\par}
\end{snapbox}
\end{minipage}\hfill
\begin{minipage}[t]{0.498\textwidth}
\begin{snapbox}[equal height group=fig4eq]{Karpathy Wiki (gpt-4.1-mini) -- Failure}
\footnotesize
\textbf{1.\,\stagepass\,Encoding}: all facts encoded\\[2pt]
{\tiny\hangindent=1em\hangafter=1
\texttt{daily/2023-03-13.md}: recipient is \hPre{Hyunwoo Nam}; \hRule{will change to James Lee if team lead changes}\par}\vspace{6pt}
\textbf{2.\,\stagepass\,Maintenance}: all facts retained\\[2pt]
{\tiny\hangindent=1em\hangafter=1
\texttt{daily/2023-03-17.md}: \hChg{Jihoon Ryu is the new team lead}\par}\vspace{6pt}
\textbf{3.\,\stagefail\,Retrieval}: change event not surfaced\\[2pt]
{\tiny\hangindent=1em\hangafter=1
\texttt{daily/2023-03-13.md}: \hPre{Hyunwoo Nam}; \hRule{\ldots if team lead changes}\\
\texttt{[daily/2023-03-17.md never opened]}\par}
\end{snapbox}
\end{minipage}\\[1pt]

\begin{snapbox}{MD-flat (Opus 4.7) -- Success}
\footnotesize
\textbf{1.\,\stagepass\,Encoding}: all facts encoded\\[2pt]
{\tiny\hangindent=1em\hangafter=1
\texttt{[2023/03/13]} Weekly report recipient is \hPre{Hyunwoo Nam} (lead = Seokjin Kang).\\
\texttt{[2023/03/13]} \hRule{Contingency: if team lead changes, recipient will be James Lee.}\par}\vspace{6pt}
\textbf{2.\,\stagepass\,Maintenance}: propagated value retained\\[2pt]
{\tiny\hangindent=1em\hangafter=1
\texttt{[2023/03/17]} \hChg{Jihoon Ryu is the new team lead.}\\
\texttt{[2023/03/17]} Per contingency, weekly report recipient is now \hGold{James Lee} (team lead changed).\par}\vspace{6pt}
\textbf{3.\,\stagepass\,Retrieval}: propagated value surfaced\\[2pt]
{\tiny\hangindent=1em\hangafter=1
- Per contingency, weekly report recipient is now \hGold{James Lee}.\par}
\end{snapbox}

\caption{State of two failing systems (Graphiti, Karpathy Wiki) and the closure case (MD-flat $\times$ Opus 4.7) on episode sw\_033, traced across encoding, maintenance, and retrieval. All six main-table systems return the pre-change value (Hyunwoo Nam); only MD-flat $\times$ Opus 4.7 propagates to James Lee. Mem0, MD-flat (gpt-4.1-mini), BM25, and text-embedding-3-small follow the same template in \cref{app:cascade-snapshots-extra}. Interactive trajectories for all six systems are browsable on the project page: \url{https://seokwonjung-jay.github.io/meme-eval/}.}
\label{fig:memory-snapshots}
\end{figure}

\begin{figure}[t]
\centering
\includegraphics[width=\linewidth]{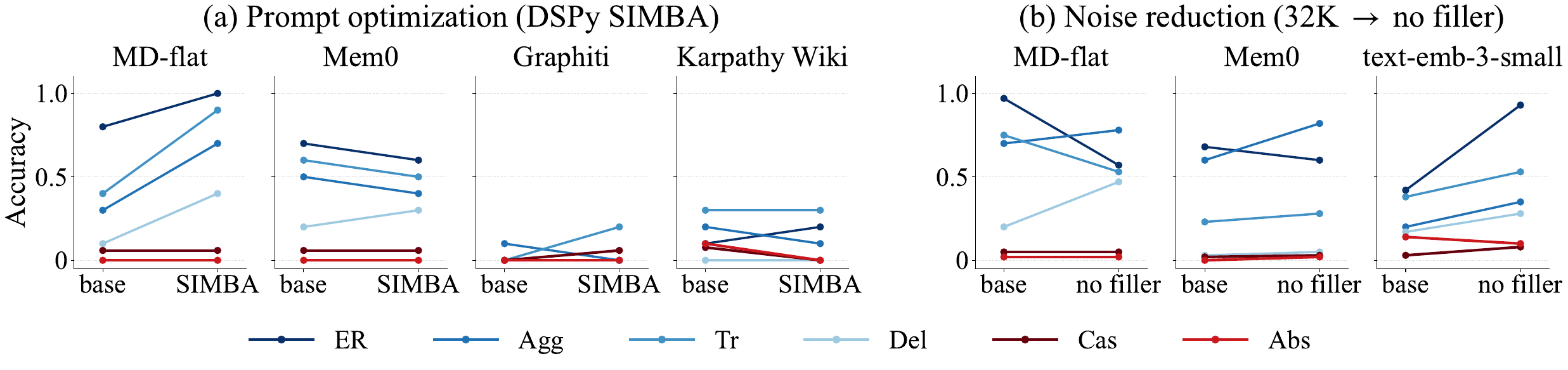}
\caption{Two interventions external to the memory architecture: (a) prompt optimization (DSPy SIMBA, 10-episode test set; 4 prompt-tunable systems), (b) noise reduction (32K filler $\to$ no filler, 40-episode subset; 3 representative systems). Cas/Abs (red lines) stay near the floor under both. Full three-condition noise sweep in \cref{fig:noise-detail}.}
\label{fig:prompt_ablation}
\end{figure}

\textbf{In-context wins on accuracy, memory wins on cost-efficiency at scale.} In-context queries on gpt-4.1-mini reach Overall 0.41, outperforming five of the six memory systems. However, in-context's per-query inference cost (\$0.16/ep) exceeds most memory systems (\$0.00--\$0.04/ep for raw retrieval, Mem0, MD-flat), so memory systems become more cost-efficient as query volume grows.

\subsection{How dependency reasoning fails}
\label{sec:where-loss}

\cref{fig:memory-snapshots} traces a representative Cascade episode through Graphiti and Karpathy Wiki, illustrating two distinct retrieval failure mechanisms. Graphiti encodes the conditional rule, the pre-change value, and the change event as edges; at query time, however, its graph search surfaces only the rule and the pre-change value, while the change-event edge falls below the retrieved top-$k$. Karpathy retains the change event in its daily log, but the query agent answers from the concept article holding the rule and the pre-change value, without drawing on the daily log that contains the change event. The remaining four systems split into two failure modes (per-system traces in \cref{app:cascade-snapshots-extra}). BM25 and MD-flat (gpt-4.1-mini) are retrieval failures: the change event falls below the top-$k$ for BM25, and MD-flat's retrieval step reads its memory file but does not extract the change entry. For text-embedding-3-small and Mem0, the change event is in the retrieved context but the answering LLM still reports the pre-change value, an answering failure.

\begin{table}[t]
\centering
\caption{Two intervention sweeps: (a) top-$k$ retrieval depth on raw retrieval and Mem0 (single-seed, 40-episode subset); (b) answering-LLM swap (gpt-4.1-mini $\to$ Sonnet 4) on all six main-table systems (100 episodes). Full per-task answer-LLM numbers in \cref{tab:answer-llm-detail}.}
\label{tab:interventions}

\begin{subtable}[t]{0.40\textwidth}
\centering
\caption*{(a) Top-$k$ retrieval depth}
\label{tab:topk-sweep}
\footnotesize
\renewcommand{\arraystretch}{0.85}
\setlength{\tabcolsep}{6pt}
\begin{tabular}{llcc}
\toprule
\textbf{System} & \textbf{top-$k$} & \textbf{Cas} & \textbf{Abs} \\
\midrule
BM25             &  5 & 0.02 & 0.08 \\
                 & 10 & 0.00 & 0.16 \\
                 & 20 & 0.02 & 0.22 \\
                 & 40 & 0.02 & 0.18 \\
\midrule
text-emb-3-small &  5 & 0.02 & 0.10 \\
                 & 10 & 0.02 & 0.18 \\
                 & 20 & 0.00 & \textbf{0.24} \\
                 & 40 & 0.00 & 0.16 \\
\midrule
Mem0             &  5 & 0.00 & 0.04 \\
                 & 10 & 0.00 & 0.02 \\
                 & 20 & 0.02 & 0.02 \\
                 & 40 & 0.00 & 0.02 \\
\bottomrule
\end{tabular}
\end{subtable}\hfill
\begin{subtable}[t]{0.55\textwidth}
\centering
\caption*{(b) Answering-LLM swap}
\label{tab:answer-llm-summary}
\footnotesize
\renewcommand{\arraystretch}{0.85}
\setlength{\tabcolsep}{4pt}
\begin{tabular}{llcc}
\toprule
\textbf{System} & \textbf{Answer LLM} & \textbf{Cas} & \textbf{Abs} \\
\midrule
BM25                   & gpt-4.1-mini & 0.02 & 0.00 \\
                       & Sonnet 4     & 0.01 & \textbf{0.12} \\
\midrule
text-emb-3-small       & gpt-4.1-mini & 0.04 & 0.00 \\
                       & Sonnet 4     & 0.03 & \textbf{0.12} \\
\midrule
Mem0                   & gpt-4.1-mini & 0.03 & 0.00 \\
                       & Sonnet 4     & 0.01 & 0.00 \\
\midrule
Graphiti               & gpt-4.1-mini & 0.02 & 0.01 \\
                       & Sonnet 4     & 0.04 & 0.00 \\
\midrule
MD-flat                & gpt-4.1-mini & \textbf{0.05} & 0.06 \\
                       & Sonnet 4     & \textbf{0.05} & 0.06 \\
\midrule
Karpathy Wiki          & gpt-4.1-mini & 0.01 & 0.03 \\
                       & Sonnet 4     & 0.01 & 0.03 \\
\bottomrule
\end{tabular}
\end{subtable}
\end{table}

\subsection{Closing the gap without changing the architecture}
\label{sec:no-closure}

\Cref{sec:where-loss} traces the gap to two stages: retrieval for BM25, MD-flat, Graphiti, and Karpathy Wiki, and answering for text-embedding-3-small and Mem0. We now test whether five interventions can close it without changing the memory architecture: prompt optimization, increased retrieval depth, a stronger answering LLM, reduced filler noise, and a stronger internal LLM. We then ask whether the answering-stage failures can be recovered by a task-specific answer prompt. Except for the answering-LLM swap and the task-specific prompt analysis, all ablations use Sonnet 4 as the answering LLM to isolate memory-system effects from the answer LLM's reasoning ceiling.

\textbf{Prompt optimization does not close the gap.} We applied DSPy SIMBA \cite{khattab2024dspy}, an append-only prompt optimizer, to MD-flat, Mem0, Graphiti, and Karpathy Wiki, optimizing each system's ingest and retrieve prompts (single-seed run on a 10-episode SIMBA test set; details in \cref{app:ablation}). Across all four systems, Cascade and Absence remain at or near the floor (\cref{fig:prompt_ablation}; SIMBA configuration in \cref{tab:ablation-config}, MD-flat multi-seed statistics in \cref{tab:simba-stability}). For three systems (MD-flat, Graphiti, Karpathy Wiki), the winning candidate appended advice explicitly targeting dependency failure modes (verbatim in \cref{app:ablation-prompts}); for Mem0, the winning candidate was the library's default extract prompt unchanged. Cas/Abs stays at the floor across all four, so better ingestion and retrieval instructions alone do not repair the gap.

\textbf{Increased retrieval depth does not help on Cascade.} For BM25, text-embedding-3-small, and Mem0, we sweep top-$k$ across $\{5, 10, 20, 40\}$ on a 40-episode subset to test whether dependency evidence is simply buried below the cutoff. Cascade remains near zero at every $k$ value across all three systems (\cref{tab:topk-sweep}). Absence on the raw-retrieval systems rises with $k$, peaking at $k{=}20$ (BM25 0.22, dense 0.24) before declining. Per-failure inspection (\cref{app:retrieval-vs-reasoning}) shows that at $k{=}20$ and $k{=}40$ both the rule and the change session are already in the retrieval context for 45\% of Cascade failures and 84\% of Absence failures. Deeper retrieval thus saturates against an answering-side bottleneck on Absence and a roughly even split on Cascade. Mem0 stays at the floor for both Cas and Abs at every $k$.

\begin{table*}[t]
\centering
\caption{Internal-LLM swap on the three systems with internal LLMs (20-episode subset, Sonnet 4 answer). \textbf{Bold} marks the best per accuracy column. Cost columns mirror \cref{tab:main_results}: \emph{Ingest} is the internal-LLM ingestion cost, \emph{Inference} is retrieval plus the Sonnet 4 answer call.}
\label{tab:llm-ablation}
\vspace{8pt}
\footnotesize
\renewcommand{\arraystretch}{0.85}
\setlength{\tabcolsep}{4pt}
\begin{tabular*}{\textwidth}{@{\extracolsep{\fill}}ll*{7}{c}*{2}{c}@{}}
\toprule
& & \multicolumn{7}{c}{\textbf{Accuracy}} & \multicolumn{2}{c}{\textbf{Cost \$/ep}} \\
\cline{3-9}\cline{10-11}
\addlinespace[3pt]
\textbf{System} & \textbf{Internal LLM} & \textbf{ER} & \textbf{Agg} & \textbf{Tr} & \textbf{Del} & \textbf{Cas} & \textbf{Abs} & \textbf{Overall} & \textbf{Ingest} & \textbf{Inference} \\
\midrule
\addlinespace[3pt]
Mem0     & gpt-4.1-mini & 0.70           & 0.60          & 0.10          & 0.05           & 0.00           & 0.00           & 0.24            & \$0.10  & \$0.04 \\
         & gpt-5        & \textbf{1.00} & 0.75          & 0.35          & 0.10          & 0.00           & 0.00           & 0.37           & \$0.34  & \$0.05 \\
         & GLM-5.1      & 0.85           & 0.65          & 0.30          & 0.15          & 0.06           & 0.00           & 0.34           & \$0.12  & \$0.04 \\
         & Opus 4.7     & 0.95           & 0.60          & 0.35          & 0.05           & 0.03           & 0.00           & 0.33           & \$2.00  & \$0.04 \\
\midrule
\addlinespace[3pt]
Graphiti & gpt-4.1-mini & 0.00            & 0.05           & 0.00           & 0.05           & 0.03           & 0.00           & 0.02           & \$0.55  & \$0.02 \\
         & gpt-5        & 0.00            & 0.00           & 0.00           & 0.05           & 0.00           & 0.00           & 0.01           & \$1.66  & \$0.01 \\
         & GLM-5.1      & 0.10           & 0.00           & 0.00           & 0.05           & 0.03           & 0.00           & 0.03           & \$0.65  & \$0.01 \\
         & Opus 4.7     & 0.00            & 0.00           & 0.00           & 0.10          & 0.00           & 0.04           & 0.02           & \$12.37 & \$0.02 \\
\midrule
\addlinespace[3pt]
MD-flat  & gpt-4.1-mini & 0.95           & 0.65          & \textbf{0.70} & 0.20          & 0.00           & 0.08           & 0.43            & \$0.04  & \$0.02 \\
         & gpt-5        & 0.25           & 0.20          & 0.15          & 0.35          & 0.00           & 0.28          & 0.20           & \$0.23  & \$0.02 \\
         & GLM-5.1      & 0.90           & 0.45          & 0.25          & 0.70          & 0.13          & 0.12          & 0.42           & \$0.09  & \$0.02 \\
         & Opus 4.7     & 0.60           & \textbf{0.80} & 0.20          & \textbf{0.80} & \textbf{0.32} & \textbf{0.60} & \textbf{0.55} & \$1.29  & \$0.23 \\
\bottomrule
\end{tabular*}
\end{table*}

\textbf{A stronger answering LLM does not consistently help.} We replace the answering LLM (gpt-4.1-mini $\to$ Claude Sonnet 4) on all six main-table systems and 100 episodes (\cref{tab:answer-llm-summary}; full per-task breakdown in \cref{tab:answer-llm-detail}). BM25 and text-embedding-3-small show small Absence gains ($0.00 \to 0.12$ each), but Cascade does not improve on any system (averages $0.03 \to 0.02$). A capable answering LLM cannot reconstruct dependencies that the memory layer never preserved.

\textbf{Reducing noise does not close the gap.} We compare the default 32K-filler condition against a no-filler condition on the highest-overall system within each paradigm: MD-flat for file-based, Mem0 for LLM-processed, and text-embedding-3-small for raw retrieval (per \cref{tab:main_results}). Cascade and Absence remain at or near the floor in both conditions (\cref{fig:prompt_ablation}b). The full three-condition sweep (no filler, 32K, 128K) is in \cref{fig:noise-detail}.

\textbf{Even the strongest internal LLM closes the gap only partially, at high cost.} Swapping the internal LLM (gpt-4.1-mini $\to$ gpt-5, GLM-5.1, or Claude Opus 4.7) on Graphiti, Mem0, and MD-flat over a 20-episode subset leaves Cascade and Absence near the floor for Mem0 and Graphiti. On MD-flat, GLM-5.1 lifts both only modestly (0.13 and 0.12), while gpt-5 raises Absence to 0.28 but leaves Cascade at zero. The largest gain is MD-flat with Opus 4.7, which reaches Cascade 0.32 and Absence 0.60 (\cref{tab:llm-ablation}). The mechanism is illustrated in the MD-flat $\times$ Opus 4.7 panel of \cref{fig:memory-snapshots} and detailed in \cref{qual:mdflat-opus-contingency}. At ingest, Opus writes the current value and an explicit contingency entry naming the parent. When the change later arrives, it scans for dependent contingency entries and writes the propagated value in place. The propagated fact is a stand-alone declarative that the retriever ranks at the top of the next query and the answering LLM reads directly, bypassing both the retrieval and answering failures of \cref{sec:where-loss}. The same Opus does not help Mem0 or Graphiti (Cas 0.00--0.03); their fact decomposition and triple extraction strip the contingency wording, so closure needs both a capable internal LLM and a substrate that preserves it. MD-flat $\times$ Opus runs at ${\sim}23\times$ the gpt-4.1-mini baseline (\cref{tab:llm-cost}) and degrades Exact Recall and Tracking. Even at this cost, both tasks remain below the gold-facts ceiling of the same answering LLM (Cascade 0.60, Absence 0.80). A three-seed re-run of this ablation preserves this ordering (\cref{app:internal-llm-seeds}).

\textbf{Task-specific prompting recovers answering failures but not retrieval failures.} Finally, we test whether the failures of \cref{sec:where-loss} can be recovered at the answering step alone. On a 20-episode subset (\cref{app:multistep}), we replace the single-step answer prompt with a task-specific one that walks the answering LLM through the dependency: identify the rule, check whether the source changed, then state the value. For text-embedding-3-small and Mem0, whose failures \cref{sec:where-loss} traces to the answering stage, Cascade and Absence rise to 0.34 and 0.42 and to 0.12 and 0.38, respectively. The retrieval failures do not recover: MD-flat and Karpathy Wiki stay at or near zero, and BM25 and Graphiti gain at most 0.16. This is a diagnostic, not a practical fix: the prompt hands the answering LLM the solution procedure for exactly the reasoning the benchmark evaluates, and applied to every question it lowers static accuracy (Exact Recall falls from 1.00 to 0.55 for BM25 and from 0.95 to 0.55 for text-embedding-3-small).

\section{Conclusion}

Long-running agentic systems demand more from memory than retrieval over a fact store. As user-stated context evolves, an agent's memory must propagate updates through related facts and signal uncertainty when no resolution rule is available. MEME tests this through Cascade and Absence, two dependency-reasoning task types that no prior benchmark covers.

{Under default configurations, none of the evaluated memory systems solves dependency reasoning. Most systems store both the dependency rule and the change event, yet fail to surface the change at retrieval or, when it is retrieved, to use it when answering. A task-specific answer prompt recovers part of the answering failures but little of the retrieval failures. The largest gain comes from a frontier internal LLM (Opus 4.7) on a file-based substrate, which writes the propagated value into the store at ingest. Even so, it closes the gap only partially at ${\sim}23\times$ the baseline token cost, remaining below the gold-facts ceiling on both Cascade and Absence.

Our per-stage diagnosis points to three capabilities a memory system needs, one per stage. At ingest, dependencies should be stored as first-class information rather than discarded as incidental. At maintenance, an update should propagate through stored dependencies when it arrives, since deferring that reasoning to query time leaves the relevant links to be found among noise. At retrieval, the system should surface dependent facts that share no surface overlap with the query, such as a commute time when the user asks about a workplace. We leave the architecture open and offer MEME as a diagnostic for tracking progress toward it.}

\section{Limitations}
Several limitations bound our findings. The dataset uses two hand-crafted knowledge graphs (Personal Life, Software Project) authored by us. Broader domains and crowd-sourced graphs would test generalization further. The structural slices in \cref{sec:main-results} come from these two graphs, not from independently built ones. Dialogues are LLM-generated rather than collected from real users, which constrains realism. We evaluate on 100 episodes with $\sim$35K-token contexts. Longer 128K-token contexts are tested only on a 40-episode subset of three systems (\cref{fig:noise-detail}), and larger samples may surface different patterns. Most ablation studies (prompt optimization, top-$k$ sweep, noise ablation, internal-LLM swap) run on 10--40 episode subsets and cover only some of the six memory systems due to compute cost. We have not tested episodes that omit the dependency entirely, in which a system would have to infer it from world knowledge. All data is English.

\begin{ack}
This work was partly supported by NAVER Corporation.

This work was partly supported by Institute for Information \& communications Technology Planning \& Evaluation(IITP) grant funded by the Korea government(MSIT) (RS-2019-II190075, Artificial Intelligence Graduate School Program(KAIST)).

This work was partly supported by the National Research Foundation of Korea (NRF) grant funded by the Korean government (MSIT) in 2026 (Grant No. RS-2026-25612350, Project Title: Development of an End-to-End Design and Validation Framework for Mathematically Principled Next-Generation AI Architectures and Learning Algorithms).

AR and AU acknowledge support from the German Federal Ministry of Education and Research (BMBF): T\"ubingen AI Center, FKZ: 01IS18039A. AR and AU additionally thank the International Max Planck Research School for Intelligent Systems (IMPRS-IS).

The authors declare no competing interests.
\end{ack}

\newpage
\appendix

\section{Operational Costs}
\label{app:costs}

We report token usage at three pipeline stages: \textbf{Ingest} (LLM calls made when storing new sessions into memory), \textbf{Retrieve} (LLM calls fetching context for a question), and \textbf{Answer} (the final answering LLM call that generates the user-facing response from retrieved context). \textbf{In} and \textbf{Out} columns separate prompt and completion tokens. Pricing per 1M tokens (input/output): gpt-4.1-mini \$0.40/\$1.60, GLM-5.1 \$0.50/\$1.50, gpt-5 \$1.25/\$10, Opus 4.7 \$5/\$25, Sonnet 4 \$3/\$15. \cref{tab:costs} reports the gpt-4.1-mini baseline configuration used in the main results (\cref{tab:main_results}); \cref{tab:llm-cost} reports the internal-LLM ablation (3 systems $\times$ 4 LLMs, 20-episode subset; answering LLM held at Sonnet 4). The partial Cascade/Absence closure observed for MD-flat $\times$ Opus 4.7 (\cref{tab:llm-ablation}) comes at $\sim$23$\times$ the baseline cost.

\begin{table}[h]
\centering
\caption{Per-episode token usage and nominal cost for the baseline configuration; all systems run end-to-end on gpt-4.1-mini. $^1$Embedding API cost ({$\sim$}\$0.0007/ep) is negligible and omitted; Mem0 and Graphiti similarly use embeddings internally, also excluded. $^2$Karpathy Wiki uses its native query pipeline for retrieval and answer generation; the cost lives in the Retrieve column.}
\label{tab:costs}
\footnotesize
\renewcommand{\arraystretch}{0.85}
\begin{tabular*}{\textwidth}{@{\extracolsep{\fill}}l*{7}{r}@{}}
\toprule
& \multicolumn{6}{c}{\textbf{Tokens / episode (mean)}} & \\
\cmidrule(lr){2-7}
\textbf{System} & \textbf{Ingest in} & \textbf{Ingest out} & \textbf{Retrv in} & \textbf{Retrv out} & \textbf{Ans in} & \textbf{Ans out} & \textbf{\$/ep} \\
\midrule
BM25                       & 0         & 0      & 0       & 0     & 93,926 & 276   & \$0.038 \\
text-embedding-3-small$^1$ & 0         & 0      & 0       & 0     & 76,821 & 272   & \$0.031 \\
Mem0                       & 223,522   & 8,703  & 0       & 0     & 11,449 & 414   & \$0.109 \\
Graphiti                   & 1,297,456 & 22,146 & 0       & 0     & 3,872  & 239   & \$0.556 \\
MD-flat                    & 89,203    & 4,990  & 25,211  & 760   & 1,532  & 346   & \$0.056 \\
Karpathy Wiki$^2$          & 2,660,556 & 67,487 & 531,555 & 2,248 & 0      & 0     & \$1.388 \\
\bottomrule
\end{tabular*}
\end{table}

\begin{table}[h]
\centering
\caption{Per-episode token usage and nominal cost for the internal-LLM ablation. Three systems $\times$ four internal LLMs on a 20-episode subset; answering LLM held at Sonnet 4. The Ans columns report Sonnet 4 answer-call tokens, included in \$/ep.}
\label{tab:llm-cost}
\footnotesize
\renewcommand{\arraystretch}{0.85}
\setlength{\tabcolsep}{3pt}
\begin{tabular*}{\textwidth}{@{\extracolsep{\fill}}ll*{6}{r}rr@{}}
\toprule
& & \multicolumn{6}{c}{\textbf{Tokens / episode (mean)}} & & \\
\cmidrule(lr){3-8}
\textbf{System} & \textbf{Internal LLM} & \textbf{Ingest in} & \textbf{Ingest out} & \textbf{Retrv in} & \textbf{Retrv out} & \textbf{Ans in} & \textbf{Ans out} & \textbf{\$/ep} & \textbf{vs base} \\
\midrule
Mem0     & gpt-4.1-mini & 223,522   & 8,703  & 0      & 0     & 11,449 & 414  & \$0.144 & 1$\times$ \\
         & gpt-5        & 229,781   & 5,611  & 0      & 0     & 13,723 & 515  & \$0.392 & 2.7$\times$ \\
         & GLM-5.1      & 230,574   & 5,994  & 0      & 0     & 10,952 & 478  & \$0.164 & 1.1$\times$ \\
         & Opus 4.7     & 360,408   & 7,912  & 0      & 0     & 10,948 & 467  & \$2.040 & 14.2$\times$ \\
\midrule
Graphiti & gpt-4.1-mini & 1,297,456 & 22,146 & 0      & 0     & 3,872  & 239  & \$0.570 & 1$\times$ \\
         & gpt-5        & 1,202,858 & 15,523 & 0      & 0     & 3,367  & 161  & \$1.671 & 2.9$\times$ \\
         & GLM-5.1      & 1,227,645 & 24,749 & 0      & 0     & 3,290  & 177  & \$0.663 & 1.2$\times$ \\
         & Opus 4.7     & 2,201,282 & 54,728 & 0      & 0     & 3,646  & 219  & \$12.39 & 21.7$\times$ \\
\midrule
MD-flat  & gpt-4.1-mini & 89,203    & 4,990  & 25,211 & 760   & 1,532  & 346  & \$0.065 & 1$\times$ \\
         & gpt-5        & 146,904   & 4,174  & 9,559  & 667   & 1,194  & 211  & \$0.251 & 3.9$\times$ \\
         & GLM-5.1      & 162,081   & 6,779  & 21,483 & 818   & 1,636  & 378  & \$0.114 & 1.8$\times$ \\
         & Opus 4.7     & 222,802   & 7,018  & 34,770 & 1,856 & 1,651  & 343  & \textbf{\$1.520} & \textbf{23.5$\times$} \\
\bottomrule
\end{tabular*}
\end{table}

\section{Dataset Construction Details}
\label{app:dataset}

\begin{table}[h]
\centering
\footnotesize
\renewcommand{\arraystretch}{0.85}
\caption{MEME dataset statistics overview, per domain. (a) Knowledge graph structure: each domain is built on a hand-crafted DAG with the entity-type breakdown shown. (b) Episode generation and evaluation statistics.}
\label{tab:dataset_stats_detail}

\begin{subtable}[t]{0.48\textwidth}
\centering
\caption*{(a) Knowledge graph}
\setlength{\tabcolsep}{4pt}
\begin{tabular}{lcc}
\toprule
 & \makecell{\textbf{Personal}\\\textbf{Life}} & \makecell{\textbf{Software}\\\textbf{Project}} \\
\midrule
Total entities      & 39  & 51 \\
~~Root              & 5   & 5  \\
~~Middle            & 5   & 6  \\
~~Leaf              & 19  & 21 \\
~~Orphan            & 10  & 19 \\
Dependency edges    & 34  & 27 \\
\bottomrule
\end{tabular}
\end{subtable}\hfill
\begin{subtable}[t]{0.48\textwidth}
\centering
\caption*{(b) Episodes and evaluation}
\setlength{\tabcolsep}{4pt}
\begin{tabular}{lcc}
\toprule
 & \makecell{\textbf{Personal}\\\textbf{Life}} & \makecell{\textbf{Software}\\\textbf{Project}} \\
\midrule
Number of episodes        & 50   & 50   \\
Evidence sessions / epi.  & 5    & 5    \\
Filler sessions / epi.    & 18   & 14.9 \\
Filler tokens / epi.      & 32K  & 32K  \\
Total questions           & 328  & 356  \\
\bottomrule
\end{tabular}
\end{subtable}
\end{table}

\subsection{Knowledge Graph Structure}
\label{app:kg}

Each domain's knowledge graph $\mathcal{G} = (V, E, \mathcal{P})$ is hand-crafted by the authors once and reused across all episodes. The structure summary appears in \cref{tab:dataset_stats_detail}(a).

\textbf{Personal Life domain.}
The personal life graph contains 39 entities organized into 9 categories (\cref{tab:pl-entities}).
The 5 rotation root entities are: residence\_location, employer, health\_condition, relationship\_status, and school.
Dependencies follow 11 semantic patterns: proximity (5 edges), company\_policy (5), life\_event (6), medical\_causation (5), priority\_shift (4), infrastructure (3), distance (2), activity\_facility (1), schedule\_conflict (1), preference (1), and curriculum (1). Each pattern provides a natural-language template that binds a target fact to its source entity with an explicit conditional dependency clause (e.g., the proximity pattern generates sentences of the form ``\{target\_fact\} --- this depends on where \{source\_entity\_phrase\}; if I move, this would change'').

\begin{table}[h]
  \caption{Personal Life domain: entity categories.}
  \label{tab:pl-entities}
  \centering
  \footnotesize
\renewcommand{\arraystretch}{0.85}
  \begin{tabular}{@{}l l@{}}
    \toprule
    \textbf{Category} & \textbf{Entities} \\
    \midrule
    Living \& Commute & residence\_location, housing\_type, living\_arrangement, \\
                      & commute\_method, commute\_duration, vehicle \\
    Work              & employer, job\_title, work\_location, work\_schedule, work\_project \\
    Health \& Fitness & health\_condition, medication, exercise\_routine, \\
                      & fitness\_facility, sleep\_pattern \\
    Food \& Diet      & diet\_preference, dietary\_restriction, favorite\_restaurant, \\
                      & food\_allergy \\
    Family            & relationship\_status, partner, family\_event \\
    Finance           & financial\_goal, insurance, subscription\_service, planned\_purchase \\
    Schedule          & travel\_plan, regular\_appointment, upcoming\_event \\
    Hobbies           & hobby, sports, club\_membership, media\_consumption, \\
                      & social\_gathering \\
    Miscellaneous     & pet, school, skill\_acquisition, life\_philosophy \\
    \bottomrule
  \end{tabular}
\end{table}

\textbf{Software Project domain.}
The software project graph contains 51 entities organized into 6 categories (\cref{tab:sw-entities}).
The 5 root entities are: framework, database, deploy\_target, team\_lead, and auth\_provider.
Dependencies follow 6 semantic patterns: tech\_compatibility (4 edges), derived\_config (6), data\_layer (4), infra\_coupling (5), team\_assignment (4), and auth\_coupling (4).
All entity values use fictitious names manually authored by the authors (e.g., Veltrion for framework, Crysthene DB for database, Narvex Cloud for deploy target, Verithos Auth for auth provider), chosen to sound plausible within their respective domains while avoiding collision with real products. This prevents parametric knowledge contamination, as the LLM cannot have encountered these names during pre-training. The same value pool is reused across all 100 episodes for consistency.

\begin{table}[h]
  \caption{Software Project domain: entity categories.}
  \label{tab:sw-entities}
  \centering
  \footnotesize
\renewcommand{\arraystretch}{0.85}
  \begin{tabular}{@{}l l@{}}
    \toprule
    \textbf{Category} & \textbf{Entities} \\
    \midrule
    Framework \& Build & framework, build\_tool, build\_command, test\_framework, \\
                       & test\_command, project\_structure, dev\_server\_command \\
    Data Layer         & database, orm\_library, model\_syntax, migration\_tool, \\
                       & connection\_string, backup\_schedule \\
    Deployment \& Infra & deploy\_target, ci\_config, deploy\_command, monitoring\_url, \\
                        & staging\_url, docker\_image, dockerfile\_path, log\_drain\_endpoint \\
    Team \& Process    & team\_lead, code\_reviewer, escalation\_contact, \\
                       & approval\_authority, weekly\_report\_recipient \\
    Auth \& Security   & auth\_provider, auth\_method, middleware\_config, \\
                       & token\_format, login\_endpoint, user\_session\_ttl \\
    Miscellaneous      & error\_log, sprint\_deadline, secret\_manager, package\_manager, \\
                       & branch\_strategy, design\_system, meeting\_day, slack\_channel, \\
                       & standup\_format, default\_branch, release\_cadence, \\
                       & test\_coverage\_target, code\_review\_policy, oncall\_rotation, \\
                       & alert\_channel, incident\_runbook\_url, documentation\_url, \\
                       & onboarding\_guide, changelog\_location \\
    \bottomrule
  \end{tabular}
\end{table}

\subsection{Entity Values and Episode Generation}
\label{app:value-pools}

Each entity $v$ has a finite value pool $\mathcal{P}(v)$ from which values are sampled. Pool sizes range from 5 to 15 values per entity. All pools are manually curated to ensure values are mutually distinguishable and semantically appropriate for their entity type. Root entities have pools of 6--12 values to support diverse episode generation across 100 episodes without excessive repetition. The complete value pools are provided in the released dataset files.

Null-like values (e.g., ``none'', ``none currently'', ``no pet'', ``not currently enrolled'') are included in some pools but are excluded from task assignment via a \texttt{NULL\_VALUES} filter. These values cannot appear as before- or after-values in cascade, update, or tracking tasks.

Episode generation follows the five-step process described in Section~3.2. A root entity is selected from the graph, cycling through all 5 roots before reuse. The cascade chain is constructed by traversing the dependency graph outward from the root. At each hop level, two types of targets are selected: one Cascade target (with a declared if-then replacement rule) and one Absence target (without a replacement value). Both 1-hop and 2-hop descendants are considered, yielding up to 4 dependency targets per episode (2 Cascade + 2 Absence). If a 2-hop chain does not exist for the selected root, the episode contains only 1-hop dependency targets.

Entities not participating in the cascade chain are assigned to stateless tasks. Exact recall selects entities explicitly marked as \texttt{exact\_recall} in the entity pool. These are entities with long, complex values that require verbatim reproduction (\texttt{life\_philosophy} for personal life, \texttt{error\_log} for software project). Deletion selects an orphan entity whose value is personally meaningful. Aggregation selects from a set of manually curated triples of semantically related but independently stored entities (e.g., hobby + sports + club\_membership for ``What do I do in my free time?''). Tracking selects from a predefined set of entities suitable for multi-value history (e.g., vehicle, media\_consumption, partner for personal life; sprint\_deadline, secret\_manager, package\_manager for software project) and assigns 3 chronologically ordered values.

Values are sampled uniformly from each entity's pool, subject to domain-specific consistency constraints applied as a post-processing pass. For dependency entities, the after-value for Cascade targets is determined by the declared if-then rule, while Absence targets have their after-value set to $\bot$ (unknown). Each generated episode must contain all 6 required task types. If any task type is missing, the entire episode is discarded and regenerated with a different root (up to 20 retries).

\subsection{Example Generated Session}
\label{app:example-session}

To illustrate the self-chat verbalization described in Section~3.2, \cref{fig:example-session-full} shows a full generated session from Episode 1 of the Personal Life domain. This session is the Fact Introduction (Part 1a) of the episode, with root entity \emph{health\_condition}. It encodes five gold facts: \emph{hobby = pottery}, \emph{vehicle = Zyvanta Sedan}, \emph{health\_condition = lactose intolerance}, \emph{sleep\_pattern = varies by shift}, and \emph{media\_consumption = The Quorath Codex (book)}. Assistant turns are abbreviated to their first sentence for space; the User LLM interleaves each fact as a natural conversational opener.

\begin{figure}[h]
\centering
\small
\fbox{\parbox{0.92\linewidth}{%
\textbf{Assistant:} ``Hi! How can I help you today?''\\[2pt]
\textbf{User:} ``My hobby is pottery. I've been getting into it more lately. Do you know any good techniques for beginners?''\\[2pt]
\textbf{Assistant:} ``One effective technique for beginners is hand-building, which includes pinch pots, coil building, and slab construction\ldots''\\[2pt]
\textbf{User:} ``I drive a Zyvanta Sedan. It's been a reliable car for me. Do you have any tips on keeping a car in good condition?''\\[2pt]
\textbf{Assistant:} ``To keep your car in good condition, regular maintenance is key: oil changes, fluids, tire pressure, and the manufacturer's maintenance schedule\ldots''\\[2pt]
\textbf{User:} ``My health condition is lactose intolerance. It can be a bit challenging at times, especially when it comes to choosing what to eat. Have you dealt with lactose intolerance before? Any tips on managing it?''\\[2pt]
\textbf{Assistant:} ``One effective approach is to explore lactose-free dairy alternatives such as almond, soy, or oat milk, and to check ingredient labels since lactose can be hidden in processed foods\ldots''\\[2pt]
\textbf{User:} ``My sleep usually varies by shift. It can be tough to maintain a consistent sleep schedule. Do you have any strategies for managing sleep when working shifts?''\\[2pt]
\textbf{Assistant:} ``Create a relaxing pre-sleep routine, use blackout curtains or earplugs to block out light and noise, and maintain a consistent sleep duration when you can\ldots''\\[2pt]
\textbf{User:} ``I've been watching a show called The Quorath Codex. It's been pretty engaging so far. Have you seen it, or do you have any recommendations for similar shows?''\\[2pt]
\textbf{Assistant:} ``If you enjoy The Quorath Codex, you might also like The Expanse, Shadow and Bone, or The Witcher\ldots''%
}}
\caption{Full generated session for Episode 1, Session 1 (Fact Introduction) of the Personal Life domain. Each gold fact is introduced by the User LLM in a natural conversational turn, followed by a topic-appropriate Assistant LLM response (abbreviated).}
\label{fig:example-session-full}
\end{figure}

\subsection{Haystack Assembly and Filler Filtering}
\label{app:haystack}

Filler sessions are drawn from two public corpora: LongMemEval-S~\cite{wu2024longmemeval} haystack sessions for Personal Life (17,034 sessions after extracting non-evidence sessions from each instance and deduplicating by session ID), and the ShareGPT coding English subset~\cite{ryokoai2023sharegpt} for Software Project (29,502 conversations). A length filter (500--5,000 estimated tokens per session) removes overly short or long fillers, yielding raw pools of 1,130 (Personal Life) and 9,135 (Software Project) sessions.

\paragraph{Filtering pipeline.}
We then filter each raw pool against that domain's gold-fact vocabulary through a four-step pipeline:
\begin{enumerate}
    \item Enumerate all possible gold-fact sentences by applying each entity's value pool to its fact template.
    \item For each gold fact, retrieve the top-$K{=}10$ candidate fillers using a hybrid of BM25 lexical scoring and \texttt{text-embedding-3-small} dense similarity.
    \item Judge each (gold fact, filler) pair with GPT-4o-mini using the prompt in \cref{app:prompts-filler}. The judge flags three conflict types: \textbf{A} (CONTRADICTION), \textbf{B} (ALTERNATIVE), and \textbf{C} (ENTITY\_CONFUSION). Fillers flagged for any type are removed from the pool.
    \item At episode assembly time, a keyword-based blocklist derived from the current episode's gold entities acts as a final safety net.
\end{enumerate}

\paragraph{Filter statistics.} \Cref{tab:filler-filter-stats} summarizes pool sizes before and after LLM-based conflict filtering. Personal Life is dominated by CONTRADICTION flags while Software Project is dominated by ALTERNATIVE flags, reflecting the different conflict profiles of everyday versus technical content.

\paragraph{License attribution.} LongMemEval~\cite{wu2024longmemeval} is released under MIT, and the ShareGPT 52K subset~\cite{ryokoai2023sharegpt} is released under CC0~1.0. We redistribute the filtered subsets used in our pipeline as \texttt{meme-benchmark/MEME-fillers} (\url{https://huggingface.co/datasets/meme-benchmark/MEME-fillers}): the structural filtering and curation are released under MIT, while the filler conversation contents pass through under their original sources' licenses. The MEME dataset of evaluation episodes is released under CC~BY~4.0.

\begin{table}[h]
\vspace{8pt}
\centering\footnotesize
\renewcommand{\arraystretch}{0.85}
\setlength{\tabcolsep}{4pt}
\begin{tabular}{lrrrr}
\toprule
Domain & Pre-filter & Post-filter & Removed & A / B / C flags \\
\midrule
Personal Life & 1,130 & 1,009 & 121 & 148 / 48 / 17 \\
Software Project & 9,135 & 9,008 & 127 & 34 / 253 / 55 \\
\bottomrule
\end{tabular}
\vspace{4pt}
\caption{Filler pool sizes before and after LLM conflict filtering. Flag counts are per (gold fact, filler) pair; a single filler may trigger multiple flags.}
\label{tab:filler-filter-stats}
\end{table}

\paragraph{Rejection examples.} \Cref{fig:filler-rejection-examples} shows one rejection example per conflict type, with the conflicting phrases in \textbf{bold} in both the gold fact and the filler excerpt.

\begin{figure}[h]
\centering
\small
\fbox{\parbox{0.92\linewidth}{%
\textbf{Type A (CONTRADICTION) --- Personal Life}\\[2pt]
\textbf{Gold:} ``My home is a \textbf{one-bedroom apartment}.''\\[2pt]
\textbf{Filler} (\texttt{filler\_6339})\textbf{:} ``\ldots We're planning to redo the yard at my \textbf{family home, which is now fully owned by me and my siblings} after finally paying off the mortgage\ldots''\\[4pt]
\rule{\linewidth}{0.4pt}\\[2pt]
\textbf{Type B (ALTERNATIVE) --- Software Project}\\[2pt]
\textbf{Gold:} ``Our build tool is \textbf{Dranith Compiler}.''\\[2pt]
\textbf{Filler} (\texttt{filler\_1096})\textbf{:} ``\ldots To \textbf{build and test software using Travis CI}, you must first select a language that is supported by Travis\ldots''\\[4pt]
\rule{\linewidth}{0.4pt}\\[2pt]
\textbf{Type C (ENTITY\_CONFUSION) --- Software Project}\\[2pt]
\textbf{Gold:} ``Our team lead is \textbf{Sarah Chen}.''\\[2pt]
\textbf{Filler} (\texttt{filler\_7586})\textbf{:} ``\ldots Reviewers: Caitlin Barrett, Jeff Kantarek, Erin Nicolle, Robert Page, Yenny Zhang, \textbf{Chris Chen}, Kelly Chang\ldots''%
}}
\caption{Rejection examples per conflict type, with conflicting phrases in \textbf{bold}. \textbf{A (CONTRADICTION)}: the family home jointly held with siblings directly contradicts the one-bedroom apartment. \textbf{B (ALTERNATIVE)}: Travis CI is a plausible alternative build tool to Dranith Compiler, even though the filler never names it in place of the gold fact. \textbf{C (ENTITY\_CONFUSION)}: Chris Chen shares a surname with the gold team lead Sarah Chen, and could be misinterpreted as the same person. Personal Life fillers are indexed in the LongMemEval-S haystack pool; Software Project fillers are indexed in the ShareGPT coding English subset.}
\label{fig:filler-rejection-examples}
\end{figure}

\paragraph{Assembly and interleaving.}
Given a total filler budget $B$ (32K tokens by default) and $N$ gaps between evidence sessions, fillers are placed into each gap in shuffled order until each gap's share of roughly $B/N$ tokens is filled. A representative 32K-token Personal Life episode consists of 20 filler and 5 evidence sessions arranged as $[F^3, E_1, F^3, E_2, F^3, E_3, F^3, E_4, F^5, E_5, F^3]$, where $F^n$ denotes $n$ consecutive fillers and $E_1$--$E_5$ denote evidence sessions (four Fact Introduction variants followed by a Change+Delete Event).

\section{Memory System Configuration}
\label{app:systems}

For each off-the-shelf system (BM25, text-embedding-3-small, Mem0, Graphiti, Karpathy Wiki), we run with the library's default configuration. Each system's defaults (top-$k$, chunk size, retrieval mix, internal-LLM logic) encode design choices that are part of the system itself; unifying them across systems would mask exactly the architectural differences MEME aims to evaluate. Parameters explicitly varied for an ablation are noted where each experiment is described. MD-flat is our own minimal-baseline implementation.

\begin{table}[H]
\centering
\setlength{\abovecaptionskip}{0pt}
\setlength{\belowcaptionskip}{0pt}
\caption{Per-system overview. \emph{Encoding}, \emph{Maintenance}, and \emph{Retrieval} characterize each system's internal handling at the corresponding pipeline stage. All systems with an internal LLM use gpt-4.1-mini; raw retrieval (BM25, text-embedding-3-small) has none. The external answering LLM is the same gpt-4.1-mini call across all systems (Karpathy Wiki fuses retrieval and answer into a single agentic loop). Text highlighting in the table body groups shared stage-level mechanisms: light green for raw-retrieval cells (no internal LLM), light blue for LLM-processed memory, and light orange for file-based tool-use. Per-system descriptions, including the storage substrate, follow below.}
\label{tab:pipeline-overview}
\footnotesize
\renewcommand{\arraystretch}{0.85}
\setlength{\tabcolsep}{4pt}
\begin{tabular*}{\linewidth}{@{\extracolsep{\fill}}llll@{}}
\toprule
\textbf{System} & \textbf{Encoding} & \textbf{Maintenance} & \textbf{Retrieval} \\
\midrule
\multicolumn{4}{l}{\textit{Raw retrieval}} \\
BM25                   & {\sethlcolor{tintraw}\hl{no LLM}}        & append-only                          & top-5 (lexical)              \\
text-embedding-3-small & {\sethlcolor{tintraw}\hl{no LLM}}        & append-only                          & top-5 (cosine)               \\
\midrule
\multicolumn{4}{l}{\textit{LLM-processed memory}} \\
Mem0     & {\sethlcolor{tintllm}\hl{LLM extracts}}   & {\sethlcolor{tintllm}\hl{LLM resolves}}      & top-20 (vector)              \\
Graphiti & {\sethlcolor{tintllm}\hl{LLM extracts}}   & {\sethlcolor{tintllm}\hl{LLM edge invalid.}} & top-10 (graph + BM25 + cosine) \\
\midrule
\multicolumn{4}{l}{\textit{File-based agents}} \\
MD-flat       & {\sethlcolor{tintfile}\hl{LLM tool-use}} & {\sethlcolor{tintfile}\hl{LLM tool-use}}   & {\sethlcolor{tintfile}\hl{LLM tool-use (read/write/append)}} \\
Karpathy Wiki & {\sethlcolor{tintfile}\hl{LLM tool-use}} & {\sethlcolor{tintfile}\hl{LLM compile}}    & {\sethlcolor{tintfile}\hl{LLM navigates \texttt{index.md}}} \\
\bottomrule
\end{tabular*}
\end{table}

\textbf{BM25.} Sparse retrieval using the bm25s library \cite{lu2024bm25s} (v0.3.2) with Lucene IDF. Each session is split into 4,096-token chunks (cl100k\_base tokenizer, respecting turn boundaries) with a \texttt{[Session: <timestamp>]} header prepended. At query time, the top-$k{=}5$ chunks are concatenated as retrieval context. No internal LLM is used; the in-memory index is rebuilt incrementally per session and reset between episodes.

\textbf{text-embedding-3-small.} Dense retrieval using OpenAI text-embedding-3-small (1,536-dim, L2-normalized cosine similarity). Chunking and retrieval follow the same protocol as BM25 (session-level, 4,096-token chunks, top-$k{=}5$). No internal LLM is used for ingestion; embedding API cost is reported in \cref{tab:costs}.

\textbf{Mem0.} Automatic fact extraction and semantic search over a Qdrant vector database. Each session transcript is passed to \texttt{Memory.add()}, which uses an internal LLM to extract discrete facts as natural-language sentences. Mem0's internal LLM logic performs conflict resolution: when a new fact contradicts an existing memory, it decides whether to ADD, UPDATE, or DELETE. Retrieval via \texttt{Memory.search()} returns the top-20 most relevant facts (library default for the Mem0 version pinned in our experiment environment; we leave \texttt{top\_k} unspecified). Each instance uses a unique collection name to prevent cross-episode contamination. Internal LLM: gpt-4.1-mini. Embedding: OpenAI text-embedding-3-small (1,536-dim, Mem0 default).

\textbf{Graphiti.} A temporal knowledge graph system storing entities and relationships in Neo4j. Each session triggers: (1) entity extraction, (2) edge extraction (triples with fact descriptions), (3) entity deduplication, (4) edge deduplication. Each edge includes temporal metadata (\texttt{valid\_at}, \texttt{invalid\_at}). Retrieval combines semantic search, BM25 keyword search, and graph traversal (top-10 results). Each episode uses an isolated \texttt{group\_id}. A key design constraint: Graphiti's extraction prompt explicitly instructs ``closely paraphrase the original source sentence(s). Do not verbatim quote the original text,'' which directly explains its near-zero Exact Recall accuracy. Internal LLM: gpt-4.1-mini (injected via OpenAIClient).

\textbf{MD-flat.} The simplest architecture: an LLM with access to a virtual file system containing a single markdown file (\texttt{memory.md}). The agent receives each conversation session as input and decides what to store, update, or delete through tool calls (\texttt{read\_file}, \texttt{write\_file}, \texttt{append\_file}, \texttt{list\_files}). The ingestion prompt deliberately avoids task-specific hints: it gives only generic instructions to update changed information and remove cancelled entries, with no mention of dependencies or conditional rules. The tool-calling loop runs for a maximum of 5 rounds per operation; in practice, most complete in 2--3 rounds. Internal LLM: gpt-4.1-mini.

\textbf{Karpathy Wiki.} An implementation of Karpathy's LLM knowledge base concept \cite{karpathy2025wiki} using the claude-memory-compiler project\footnote{\url{https://github.com/coleam00/claude-memory-compiler}} without modification to its prompts or logic. Three-stage pipeline:
\begin{itemize}
  \item \textbf{Flush (ingest):} Each session is passed to \texttt{flush.py}, which uses a single LLM call (no tool use) to extract important knowledge and append it to a daily log file.
  \item \textbf{Compile (consolidation):} After ingesting all sessions, \texttt{compile.py} processes the daily logs into structured knowledge articles organized into three categories: concepts, connections, and qa. An \texttt{index.md} file serves as a catalog. The compile operation runs with \texttt{max\_turns=30} and has access to file tools (Read, Write, Edit, Glob, Grep).
  \item \textbf{Query (retrieval + answer):} At question time, \texttt{query.py} reads \texttt{index.md} to identify relevant articles, reads them, and generates an answer in a single agentic loop (\texttt{max\_turns=15}). Karpathy Wiki uses its native query pipeline for answer generation rather than the unified gpt-4.1-mini answering LM; the cost is reflected in the Retrieve column of \cref{tab:costs}.
\end{itemize}
Each episode runs in an isolated workspace. Internal LLM: gpt-4.1-mini.

\subsection{Experiment Manifest}
\label{app:manifest}

\begin{table}[h]
\centering
\footnotesize
\renewcommand{\arraystretch}{0.9}
\caption{Configuration of every reported experiment. All runs use the 32K-filler dataset (except the noise ablation), temperature 0 (except the SIMBA prompt model, 0.9), and a GPT-4o judge. Model snapshots: \texttt{gpt-4.1-mini-2025-04-14}, \texttt{gpt-4o-2024-08-06}, \texttt{gpt-5-2025-08-07}, \texttt{claude-sonnet-4-20250514}, \texttt{claude-sonnet-4-6}, \texttt{claude-opus-4-7}, and GLM-5.1 via OpenRouter (\texttt{z-ai/glm-5.1}).}
\label{tab:manifest}
\begin{tabular*}{\columnwidth}{@{\extracolsep{\fill}}lccllc@{}}
\toprule
\textbf{Experiment} & \textbf{Episodes} & \textbf{Systems} & \textbf{Internal LLM} & \textbf{Answering LLM} & \textbf{Runs} \\
\midrule
\cref{tab:main_results} memory systems & 100 & 6 & gpt-4.1-mini & gpt-4.1-mini & 1 \\
\cref{tab:main_results} in-context & 100 & -- & -- & gpt-4.1-mini, Sonnet 4.6 & 1 \\
Gold-facts ceiling (\cref{tab:in-context-ceiling}) & 100 & -- & -- & 4 options & 1 \\
Top-$k$ sweep ($k{=}5,10,20,40$) & 40 & 3 & gpt-4.1-mini & Sonnet 4 & 1 \\
Answering-LLM swap & 100 & 6 & gpt-4.1-mini & gpt-4.1-mini $\rightarrow$ Sonnet 4 & 1 \\
Internal-LLM swap (\cref{tab:llm-ablation}) & 20 & 3 & 4 options & Sonnet 4 & 1 \\
Prompt optimization (SIMBA) & 10 + 10 & 4 & gpt-4.1-mini & Sonnet 4 & 1 \\
Noise ablation (none, 32K, 128K) & 40 & 3 & gpt-4.1-mini & Sonnet 4 & 1 \\
Repeated runs (\cref{app:sd}) & 10 & 4 & gpt-4.1-mini & Sonnet 4 & 5 \\
Implicit rules (\cref{app:implicit}) & 20 & 6 & gpt-4.1-mini & gpt-4.1-mini & 1 \\
Task-specific prompt (\cref{app:multistep}) & 20 & 6 & gpt-4.1-mini & gpt-4.1-mini & 1 \\
Seed re-run (\cref{app:internal-llm-seeds}) & 20 & 3 & 3 options & Sonnet 4.6 & 3 \\
\bottomrule
\end{tabular*}
\end{table}

\section{Prompts}
\label{app:prompts}

\subsection{Verbalization Prompts}\label{app:prompts-verbalize}

The conversation generation pipeline converts structured episode skeletons into natural multi-turn conversations using a three-stage process. First, gold fact seeds (written in first-person) are batch-converted to third-person statements by an LLM. This intermediate representation is needed because the subsequent self-chat User LLM must receive objective fact descriptions rather than first-person utterances, which it then re-verbalizes naturally. Second, a User LLM and an Assistant LLM alternate in self-chat to produce the conversation. Domain-specific third-person prompts are used for each domain:

\begin{figure}[H]
\begin{promptbox}
{\small
You are rewriting first-person fact statements into third-person statements about ``the user''.

You will receive a list of first-person fact seeds. Rewrite each into a natural, grammatically correct third-person sentence about ``the user''.

STRICT RULES:

-- Convert first-person (``I'', ``my'', ``we'', ``our'') to third-person (``the user'', ``the user's'', ``they'', ``their'')

-- Fix verb conjugation: ``I work'' $\to$ ``The user works'', ``I have'' $\to$ ``The user has''

-- Fix grammar issues: articles (a/an/the), prepositions, verb forms, capitalization

-- Fix semantic mismatches: e.g., ``watching a show called a podcast'' $\to$ ``listening to a podcast called...''

-- Keep ALL proper nouns, numbers, codes, and entity values EXACTLY as given (e.g., ``Velthari Studio'', ``10-to-7 fixed'', ``RSV-09831-THETA'' must appear verbatim)

-- Keep dependency/causal relationships intact

-- Each output must be a single self-contained sentence

-- Do NOT add information not present in the seed

-- Do NOT merge multiple seeds into one sentence

-- Do NOT change punctuation within entity values

Output format: JSON object with key ``facts'' containing an array of rewritten sentences.
}
\end{promptbox}
\caption{Third-person conversion prompt (Personal Life).}
\label{fig:prompt-thirdperson-pl}
\end{figure}

\begin{figure}[H]
\begin{promptbox}
{\small
You are rewriting first-person project statements into third-person statements about ``the developer'' or ``the team''.

You will receive a list of first-person fact seeds about a software project. Rewrite each into a natural, grammatically correct third-person sentence.

STRICT RULES:

-- Convert ``we''/``our'' to ``the team''/``the team's'', ``I'' to ``the developer''

-- Fix verb conjugation: ``We use'' $\to$ ``The team uses''

-- Keep ALL tool names, commands, URLs, connection strings, and codes EXACTLY as given (e.g., ``Veltrion'', ``karnex build --prod'', ``crysthene://prod-db:5432/aurora'' must appear verbatim)

-- Keep dependency/causal relationships intact

-- Each output must be a single self-contained sentence

-- Do NOT add information not present in the seed

-- Do NOT merge multiple seeds into one sentence

Output format: JSON object with key ``facts'' containing an array of rewritten sentences.
}
\end{promptbox}
\caption{Third-person conversion prompt (Software Project).}
\label{fig:prompt-thirdperson-sw}
\end{figure}

After third-person conversion, the User LLM receives these facts and re-verbalizes them as a first-person user in natural conversation:

\begin{figure}[H]
\begin{promptbox}
{\small
I will give you a list of personal facts. Use them to act as a normal user chatting with a chat assistant. In the chat, you may ask it to assist you with various tasks or ask it about various information. However, make sure that you convey the facts about you one by one, roughly in order. IMPORTANT: convey at most ONE fact per message. Do not combine multiple facts into a single message.

After conveying each fact, continue discussing that topic for 2--3 more exchanges before moving to the next fact. Ask follow-up questions, request more details, or explore related aspects of the topic. Not every message needs to contain a new fact.

Make sure your message is concise (1--3 sentences), since real users often do not bother to write a long message. You must simulate the tone of a neutral user and do not be overly enthusiastic, verbose, formal, or polite. Do NOT say ``thanks'', ``I will do that'', or ``is there anything else'' --- you are a user, not an assistant.

Do NOT introduce personal facts about yourself beyond the list provided. Only convey the facts given below --- do not invent new hobbies, family details, health conditions, jobs, living situations, or any other personal information. Follow-up questions should be about the TOPIC, not about yourself.

VERBALIZATION RULES (follow strictly):

1. Each fact must include its entity keyword (e.g., ``My health condition is...'', ``My financial goal is...''). The entity concept must appear in your message.

2. If a fact mentions a dependency (``depends on'', ``determined by'', ``if X changes''), you MUST preserve the full dependency language including the conditional part. The reader must understand that if the source changes, the target would change too. Do NOT weaken ``depends on X; if X changes, this would change'' to just ``tied to X'' or ``since we use X''.

3. If a fact starts with ``If'' (conditional), use ONLY ``will'' as the modal: ``If X changes, Y will be Z''. NEVER use ``would/might/probably/consider''. NEVER state it as accomplished (``I switched to Z'').

4. If a fact is marked with [CHANGED], state the new value as a definitive current fact. Do NOT hedge or speculate.

You are a HUMAN user, not an AI. Never say anything that implies you are an AI, a language model, or that you have a training data cutoff.

Never repeat a previous message verbatim. Each message must be unique.

When ALL facts have been conveyed AND you have finished discussing the last topic, generate exactly: [END]

Facts to convey: \{facts\_list\}
}
\end{promptbox}
\caption{User LLM system prompt for self-chat conversation generation.}
\label{fig:prompt-user}
\end{figure}

The Assistant LLM responds naturally without introducing unsupported assumptions:

\begin{figure}[H]
\begin{promptbox}
{\small
Respond helpfully and naturally to the user's message. Write in natural prose paragraphs --- do NOT use numbered lists, bullet points, or bold headers. Keep your response to 3--6 sentences.

IMPORTANT: You must NOT assume, guess, or mention any of the following about the user unless the user explicitly stated it in THIS conversation:

-- Family/relationships (partner, spouse, parents, siblings)

-- Living situation (city, housing type, roommates)

-- Work details (company type, industry, salary, colleagues)

-- Health/medical info (conditions, medications, diet reasons)

-- Financial details (income, savings amounts, debt)

-- Schedule/routine specifics beyond what user said

-- Hobbies, pets, vehicles, schools beyond what user said

You MAY ask clarifying questions about the topic at hand. You MAY give general advice and information.

Never say ``I don't have personal experiences'', ``As an AI'', or similar disclaimers. Respond naturally as a knowledgeable assistant.
}
\end{promptbox}
\caption{Assistant LLM system prompt for self-chat conversation generation (Personal Life).}
\label{fig:prompt-assistant}
\end{figure}

\begin{figure}[H]
\begin{promptbox}
{\small
Respond helpfully and naturally to the developer's message about their project. Write in natural prose --- do NOT use numbered lists, bullet points, or bold headers. Keep your response to 3--6 sentences.

IMPORTANT: You must NOT assume, guess, or mention any of the following unless the developer explicitly stated it in THIS conversation:

-- Tool names, frameworks, databases, or libraries not mentioned

-- URLs, connection strings, API keys, or endpoints not shared

-- Team members, leads, or contacts not introduced

-- CI/CD configs, Docker images, or deploy commands not stated

-- Branching strategies, sprint details, or meeting schedules not mentioned

You MAY ask clarifying questions about the setup. You MAY give general engineering advice.

Never say ``I don't have personal experiences'', ``As an AI'', or similar disclaimers. Respond naturally as a knowledgeable assistant.
}
\end{promptbox}
\caption{Assistant LLM system prompt for self-chat conversation generation (Software Project).}
\label{fig:prompt-assistant-sw}
\end{figure}

\subsection{Verification Prompts}\label{app:prompts-verify}

\textbf{Annotation prompt.}
After self-chat generation, an LLM annotates which user turn conveyed each fact:

\begin{figure}[H]
\begin{promptbox}
{\small
You are annotating a conversation to identify where personal facts were conveyed.

Given a list of facts (in third-person) and a conversation, determine which user turn conveys each fact. A fact is ``conveyed'' if the user's message communicates the same information, even if worded differently.

Rules:

-- Only consider USER turns (not assistant turns)

-- A fact is conveyed if the core information matches, even with different wording. E.g., Fact ``The user works at Zypherix Labs'' matches user saying ``I just started at Zypherix Labs''

-- If a fact appears in multiple turns, return the FIRST (earliest) turn

-- If a fact was NOT conveyed in any turn, set turn\_index to null

-- turn\_index is 0-based index into the conversation array (counting both user and assistant turns)

Output format: JSON object with key ``annotations'' containing an array of objects:

[\{``fact\_id'': 1, ``conveyed'': true, ``turn\_index'': 2\}, ...]
}
\end{promptbox}
\caption{Annotation verification prompt (Layer 1).}
\label{fig:prompt-annotation}
\end{figure}

\textbf{Gemini semantic audit prompt.}
A separate LLM (Gemini 2.5 Flash) performs a five-point audit:

\begin{figure}[H]
\begin{promptbox}
{\small
You are a data quality auditor for MEME, a benchmark that evaluates LLM agent memory systems.

For each gold fact, check ALL criteria below. Report ONLY failures.

1. VERBALIZATION ACCURACY

Is the gold fact's value present in the conversation (exactly or close paraphrase)?

2. IF-THEN CONDITIONAL FORM (only for facts with is\_if\_then=true)

Must contain ``if'' conditional structure. Trigger entity (dependency\_source) must be explicitly mentioned. Modal must be ``will'' (not ``would/might/probably'') $\to$ flag WEAK\_MODAL. Must NOT be stated as accomplished fact $\to$ flag CONDITIONAL\_VIOLATED.

3. ENTITY KEYWORD PRESENCE

The question asks about an entity using certain words (e.g., ``What's my financial goal?''). Those words or a natural synonym must appear in the conversation where the fact is introduced. FAIL if entity concept never mentioned (e.g., ``I'm paying off debt'' without ``financial goal'').

4. CASCADE-U DEPENDENCY (only for facts with has\_dependency=true AND is\_if\_then=false)

Is the causal link STRONGLY explicit? The message must contain language like ``depends on'', ``determined by'', or ``if X changes, Y would change''. FAIL if dependency is only implied (``since we use X'', ``tied to X'') rather than explicitly conditional (``depends on X; if X changes, this would change'').

5. GOLD ANSWER FORMAT

Does before-delete gold have ``Yes --- '' prefix? (flag PREFIX\_ISSUE)

Output JSON array of issues:

\{``entity'': ``...'', ``fact\_index'': N,

~~``check'': ``VERBALIZATION $|$ IF\_THEN $|$ KEYWORD $|$ CASCADE\_U\_DEP $|$ GOLD\_FORMAT'',

~~``severity'': ``HIGH $|$ MEDIUM $|$ LOW'', ``issue'': ``brief description''\}

Severity: HIGH = makes task unsolvable. MEDIUM = may cause false pass/fail. LOW = minor.

If ALL pass: \{``status'': ``ALL\_PASS'', ``facts\_checked'': N\}
}
\end{promptbox}
\caption{Gemini semantic audit prompt (Layer 2).}
\label{fig:prompt-gemini}
\end{figure}

\subsection{Ingestion and Retrieval Prompts (MD-flat)}\label{app:prompts-ingest}

MD-flat is the only memory system whose ingestion and retrieval prompts are designed by the benchmark authors; the other systems use their own built-in logic without modification. The ingestion prompt processes each session and updates \texttt{memory.md}; the retrieval prompt reads back from \texttt{memory.md} at query time.

\begin{figure}[H]
\begin{promptbox}
{\small
You are a personal assistant with a persistent memory file (memory.md). After each conversation, save any information the user shared that may be useful in future sessions. Keep it compact --- one fact per line with timestamp [YYYY/MM/DD].

If information has changed, update it. If something was removed or cancelled, remove the old entry.

Do NOT save conversation summaries, assistant responses, or temporary task state.

You will now receive a conversation session. Read it and update your memory accordingly.
}
\end{promptbox}
\caption{MD-flat ingestion prompt. The agent operates through a tool-calling loop (max 5 rounds) with \texttt{read\_file}, \texttt{write\_file}, \texttt{append\_file}, and \texttt{list\_files}.}
\label{fig:prompt-mdflat}
\end{figure}

\begin{figure}[H]
\begin{promptbox}
{\small
You have access to \texttt{read\_memory()}. Read your memory to find relevant facts.

Extract ONLY the facts relevant to answering the question below. Read your memory file first, then return relevant facts as-is (do not rephrase or summarize). If nothing is relevant, say `(no relevant facts)'.

Question: \{question\}
}
\end{promptbox}
\caption{MD-flat retrieval prompt. At query time the agent runs a tool-calling loop with only \texttt{read\_memory} available, extracts the relevant facts from \texttt{memory.md} verbatim, and returns them (or `(no relevant facts)') as the retrieval context for the answering LLM.}
\label{fig:prompt-mdflat-retrieve}
\end{figure}

\subsection{Answer Prompt}\label{app:prompts-answer}

All memory systems share a single unified answer prompt. After each system's \texttt{retrieve()} method returns a context string, the following prompt is used with every answering LLM; Karpathy Wiki is the exception, since it answers inside its own query agent:

\begin{figure}[H]
\begin{promptbox}
{\small
Answer the user's question based ONLY on the context provided below. If the information is not in the context, say you don't have that information. Answer with ONLY the value. Do not explain or add context.

Context: \{context\}

Question: \{question\}
}
\end{promptbox}
\caption{Unified answer prompt shared across all memory systems.}
\label{fig:prompt-answer}
\end{figure}

\subsection{Judge Prompts}\label{app:prompts-judge}

The evaluation framework uses task-specific judge prompts with GPT-4o (temperature 0). Each task type has its own prompt:

\begin{figure}[H]
\begin{promptbox}
{\small
You are a judge evaluating whether an AI agent's answer is semantically correct.

QUESTION: \{question\}

GOLD: \{gold\}

AGENT: \{agent\_answer\}

Does the agent's answer contain the correct information matching the gold answer?

Rules:

-- Focus on semantic equivalence, not exact wording

-- If the gold value is present in the agent's answer, it is correct --- regardless of any additional information, future possibilities, or extra details the agent mentions

-- ``Dentist every 6 months; dermatologist monthly (if you change residence)'' $\to$ gold is ``dentist (every 6 months)'' $\to$ YES (core answer correct, extra info is irrelevant)

-- ``40 minutes'' = ``40 min'' $\to$ YES

-- ``dentist appointment every 6 months'' = ``dentist (every 6 months)'' $\to$ YES

-- If agent says ``I don't know'' and gold is a specific value $\to$ NO

Output ONLY JSON: \{``correct'': true/false, ``reason'': ``one sentence''\}
}
\end{promptbox}
\caption{Judge prompt: Before-phase (common).}
\label{fig:prompt-judge-before}
\end{figure}

\begin{figure}[H]
\begin{promptbox}
{\small
You are a judge evaluating whether an AI agent correctly recalls a sequence of historical values for an entity that changed multiple times.

QUESTION: \{question\}

GOLD (ordered history values, comma-separated): \{gold\}

AGENT: \{agent\_answer\}

The user's information changed multiple times. The question asks for the FULL HISTORY in chronological order (earliest to latest).

Rules:

-- Agent's answer must contain ALL gold values in the EXACT order specified

-- Order matters: the values must appear earliest-to-latest as in the gold

-- Extra surrounding text is fine, but the gold sequence must be preserved

-- Missing any value $\to$ NO

-- Wrong order $\to$ NO

-- Only some values $\to$ NO

-- ``I don't know'' $\to$ NO

Output ONLY JSON: \{``correct'': true/false, ``reason'': ``one sentence''\}
}
\end{promptbox}
\caption{Judge prompt: Tracking.}
\label{fig:prompt-judge-tracking}
\end{figure}

\begin{figure}[H]
\begin{promptbox}
{\small
You are a judge evaluating whether an AI agent's answer contains specific values.

QUESTION: \{question\}

TARGET VALUES: \{gold\_values\}

AGENT: \{agent\_answer\}

For EACH target value, check whether it appears (exact or paraphrased) in the agent's answer.

Rules:

-- The value appears in the answer (exact or paraphrased) $\to$ true

-- The value is absent $\to$ false

-- ``I don't know'' or ``I don't have that information'' $\to$ all false

Output ONLY JSON: \{``results'': [\{``value'': ``...'', ``present'': true/false\}], ``reason'': ``one sentence''\}
}
\end{promptbox}
\caption{Judge prompt: Aggregation.}
\label{fig:prompt-judge-agg}
\end{figure}

\begin{figure}[H]
\begin{promptbox}
{\small
You are a judge evaluating whether an AI agent correctly recalls an EXACT value verbatim.

QUESTION: \{question\}

GOLD: \{gold\}

AGENT: \{agent\_answer\}

Does the agent's answer contain the exact value?

Rules:

-- Check if the gold value appears VERBATIM as a substring in the agent's answer

-- If the gold value is fully contained in the answer, even with extra words before/after it $\to$ YES

-- Example: gold=``OOM killed by container runtime'', agent=``OOM killed by container runtime errors'' $\to$ YES (gold is fully preserved)

-- Minor formatting differences are acceptable (e.g., extra spaces, capitalization)

-- Missing or substituted words WITHIN the gold value $\to$ NO

-- ``I don't know'' or ``I don't have that information'' $\to$ NO

-- A completely different value $\to$ NO

Output ONLY JSON: \{``correct'': true/false, ``reason'': ``one sentence''\}
}
\end{promptbox}
\caption{Judge prompt: Exact Recall. Note: implemented as deterministic substring match at runtime; this prompt serves as fallback documentation.}
\label{fig:prompt-judge-exact}
\end{figure}

\begin{figure}[H]
\begin{promptbox}
{\small
You are a judge evaluating whether an AI agent correctly handles a DELETED piece of information.

QUESTION: \{question\}

GOLD: \{gold\}

AGENT: \{agent\_answer\}

The information was explicitly deleted/removed. The agent should not return it.

Rules:

-- Agent says ``I don't have that information'' / ``no longer available'' / ``Unknown'' / ``None'' / similar WITHOUT revealing the deleted value $\to$ YES

-- Agent indicates the item no longer exists or applies WITHOUT revealing the deleted value (e.g., ``No food allergies'' when the deleted value was a specific allergy) $\to$ YES

-- Agent reveals the deleted value in any way (even while saying it was deleted/removed) $\to$ NO

-- Agent returns the deleted value as if it still exists $\to$ NO

Output ONLY JSON: \{``correct'': true/false, ``reason'': ``one sentence''\}
}
\end{promptbox}
\caption{Judge prompt: Deletion.}
\label{fig:prompt-judge-deletion}
\end{figure}

\begin{figure}[H]
\begin{promptbox}
{\small
You are a judge evaluating whether an AI agent correctly reports a CASCADE-DEPENDENT value after an upstream change.

QUESTION: \{question\}

GOLD (new cascaded value): \{gold\}

AGENT: \{agent\_answer\}

The upstream entity changed, triggering a cascade. The agent should now report ONLY the new value.

Rules:

-- Agent states the new value as the sole, definitive answer $\to$ YES

-- Agent lists BOTH old and new values together $\to$ NO

-- Agent states only the old value $\to$ NO

-- Agent lists multiple options or hedges (``might'', ``considering'') $\to$ NO

-- Agent uses future tense (``will change to X'', ``will be X'') instead of confirming as current $\to$ NO

-- ``I don't know'' $\to$ NO

Output ONLY JSON: \{``correct'': true/false, ``reason'': ``one sentence''\}
}
\end{promptbox}
\caption{Judge prompt: Cascade.}
\label{fig:prompt-judge-cascade}
\end{figure}

\begin{figure}[H]
\begin{promptbox}
{\small
You are a judge evaluating whether an AI agent appropriately handles UNCERTAINTY after an upstream entity changed.

QUESTION: \{question\}

GOLD: \{gold\}

AGENT: \{agent\_answer\}

The upstream entity changed, making the previous answer potentially invalid. The agent should express uncertainty.

Rules:

-- Agent says ``I don't know'' / ``not sure'' / ``Unknown'' / ``None'' / expresses uncertainty $\to$ YES

-- Agent acknowledges the upstream change and questions validity $\to$ YES

-- Agent confidently states the OLD value as definitive current fact $\to$ NO

-- Agent provides the old value with no hedging or uncertainty $\to$ NO

Output ONLY JSON: \{``correct'': true/false, ``reason'': ``one sentence''\}
}
\end{promptbox}
\caption{Judge prompt: Absence.}
\label{fig:prompt-judge-absence}
\end{figure}

For Aggregation and Tracking tasks, partial credit is computed in addition to binary pass/fail. Aggregation records the number of target values present out of the total. Tracking records the number of history values in correct chronological order using a position-based algorithm.

\paragraph{Judge audit examples.} We audited 144 judge decisions against author labels (\cref{sec:setup}); the judge agrees on 142. Three examples illustrate agreement on a pass, agreement on a failure, and a disagreement. \emph{Agreement, pass} (Cascade, Graphiti, sw\_010). Q: ``What's the deploy command?'' Gold: ``\texttt{thandrel ship -{}-channel beta}.'' Answer: ``\texttt{thandrel ship -{}-channel beta}.'' \emph{Agreement, fail} (Cascade, MD-flat, pl\_026). The user first says they bike to work because they live in Zaravel Heights, and that they will switch to the subway if they move. After the move, Q: ``How do I get to work?'' Gold: ``subway.'' Answer: ``You bike to work.'' \emph{Disagreement} (Tracking, MD-flat, pl\_050). Q: ``List all subscription services I've had over time, from earliest to latest.'' Gold: ``Thyranote, meal kit (Kelvimeal), Dranthazon.'' Answer: ``Thyranote, Kelvimeal, Dranthazon.'' The judge marks a fail because the ``meal kit'' label is missing; the author marks a pass.

\subsection{Filler Filtering Prompt}\label{app:prompts-filler}

\begin{figure}[H]
\begin{promptbox}
{\small
You are checking if a filler conversation conflicts with a known gold fact.

Gold fact: ``\{gold\_fact\}''

Filler conversation:

---

\{filler\}

---

Check these 3 conflict types:

A) CONTRADICTION: Does the filler directly contradict the gold fact?

B) ALTERNATIVE: Does the filler introduce a plausible alternative answer that could confuse a memory system?

C) ENTITY\_CONFUSION: Does the filler mention the same entity/topic in a confusing way?

Answer in JSON format: \{``A'': true/false, ``B'': true/false, ``C'': true/false, ``reason'': ``brief explanation or `no conflict'\,''\}
}
\end{promptbox}
\caption{Filler conflict judgment prompt (GPT-4o-mini).}
\label{fig:prompt-filler}
\end{figure}

\section{Prompt Optimization Details}
\label{app:ablation}

This section provides details on the prompt optimization experiment described in \cref{sec:no-closure}.

\subsection{Setup and Configuration}

We used DSPy SIMBA \cite{khattab2024dspy}, an append-only prompt optimizer that iterates two strategies, rule appending and few-shot demo appending, without modifying or deleting baseline instructions. SIMBA was applied to MD-flat, Mem0, Graphiti, and Karpathy Wiki. For each system, only the ingest and retrieve prompts were optimized; the answer prompt was frozen because its backing model (Claude Sonnet 4) does not support the OpenAI \texttt{seed} parameter, making noise non-suppressible for that component. Multi-seed ($N{=}5$) statistics for MD-flat are reported in \cref{app:sd-simba}; Mem0, Graphiti, and Karpathy results are single-seed (seed=7).

\begin{table}[h]
\centering
\footnotesize
\renewcommand{\arraystretch}{0.85}
\caption{Model and optimizer configuration for the prompt optimization experiment.}
\label{tab:ablation-config}
\begin{tabular}{lll}
\toprule
\textbf{Role} & \textbf{Model} & \textbf{Temp.} \\
\midrule
Task LM (ingest, retrieve) & gpt-4.1-mini & 0 \\
Answer LM & Claude Sonnet 4 & 0 \\
Prompt model (SIMBA meta) & gpt-4.1-mini & 0.9 \\
Judge & GPT-4o & 0 \\
\bottomrule
\end{tabular}
\end{table}

SIMBA was configured with \texttt{max\_steps=2}, \texttt{bsize=4}, \texttt{num\_candidates=4}, and \texttt{num\_threads=4}. The data split used seed=7 stratified sampling: 10 training episodes (5 PL + 5 SW) and 10 held-out test episodes (5 PL + 5 SW), disjoint at the episode level.

\subsection{Optimized Prompt Content}\label{app:ablation-prompts}

SIMBA returned three candidate programs per system; for each, the winning candidate appended advice blocks to the system's prompts (baseline text retained verbatim before these blocks). The appended text per system is shown below.

\paragraph{MD-flat.}
\begin{figure}[H]
\begin{promptbox}
{\small
\textbf{(ingest, appended)} If the input involves information about configuration dependencies (such as backup schedule, ORM used, or model syntax), then explicitly encode conditional knowledge linking these dependencies, e.g., if ORM changes from Vorathis ODM to Pyravar Mapper, the model syntax should update accordingly. If the input indicates deletion of an entity (like the onboarding guide), then mark the entity as deleted and respond explicitly that the information is no longer available. When partial or uncertain information is present, avoid generic `I don't know' answers, and instead provide clarifications or acknowledge the absence explicitly. Leverage exact matching and dependency chains to improve recall of critical facts such as exact error logs.

If the conversation input includes conditional or dependent information (e.g., living arrangements depend on relationship status), then you should explicitly encode the dependency chains and update or remove old facts accordingly. When a change event occurs (like marriage or dietary restriction removal), ensure to mark prior conflicting facts as outdated or deleted and represent the new facts with clear dependency references so that downstream retrieval can reason over them.
\\[4pt]
\textbf{(retrieve, appended)} When receiving queries, especially those involving multiple updates, cascading dependencies, or multi-hop reasoning, focus on retrieving all relevant facts including updated and deleted information to fully answer the question. For multi-hop queries, chain together related facts to assemble a comprehensive response. When facts have dependencies or conditionals (e.g., backup schedule depends on the current database), retrieve the full dependency chain. For deletion queries, ensure retrieval reflects removal status so final answers indicate lack of availability rather than ignoring or failing to answer.

If a query involves entities with known dependencies or recent changes (such as changed relationship status affecting living arrangements and family events), then you should retrieve not only direct facts but also the full dependency chain surrounding these facts. Prioritize the most recent or updated information and infer uncertainty when dependencies conflict or are unresolved. This strategy helps generate consistent, accurate answers, especially for cascading or multi-hop questions.
}
\end{promptbox}
\caption{MD-flat: SIMBA-appended advice blocks for the ingest and retrieve prompts.}
\label{fig:prompt-simba-mdflat}
\end{figure}

The optimized retrieve prompt contains explicit instructions to ``infer uncertainty when dependencies conflict or are unresolved,'' directly targeting the Absence task. Despite this, Absence accuracy drops to 0.00, suggesting that LLMs do not reliably internalize meta-cognitive instructions about expressing uncertainty.

\paragraph{Mem0.}
SIMBA returned the unchanged Mem0 library default (\texttt{ADDITIVE\_EXTRACTION\_PROMPT}, ${\sim}33$K chars with default examples) as the winning candidate. The two alternative candidates that did append rule blocks scored lower on training (0.509 and 0.501 vs.\ baseline 0.545), so SIMBA selected the default. Test accuracy moved $0.545 \to 0.534$ ($-1.1$pp), within the noise floor reported next.

\paragraph{Graphiti.}
SIMBA appended advice paragraphs to each of the three Graphiti prompts (\texttt{extract\_message}, \texttt{edge}, \texttt{dedupe\_nodes}). The appended texts are shown below.

\begin{figure}[H]
\begin{promptbox}
{\small
\textbf{(extract\_message, appended)} If the module receives conversation messages containing technical or configuration details such as database connection strings, ORM model syntaxes, backup schedules, or package manager names, then it should prioritize explicit and accurate extraction of these distinct factual entities. It should distinguish between different software components and configurations even if phrased similarly. Being explicit in naming and careful in identifying such entities will improve downstream knowledge integration.

If the module receives conversation messages containing explicit conditional or dependent living arrangements, then it should explicitly extract and represent all relevant entities, including conditional entities like 'sibling' tied to relationship status changes. The module should robustly disambiguate references and update entity extraction as new facts (e.g., changes in relationship status) are introduced to avoid missing or outdated entities.
\\[4pt]
\textbf{(edge, appended)} If the module receives messages and an entity list containing technical terms like connection strings, model syntax, or backup schedules, then it should extract precise fact triples that link these entities with their values and contextual metadata, including temporality and versioning if applicable. It should avoid hallucination and ensure relations reflect the actual statements in the conversation, supporting correct temporal or conditional dependencies (e.g., changes of configurations over time).

If the module receives entity lists and conversation messages with conditional or evolving personal information, then it should ensure that fact triples clearly capture these conditional relationships and that responses to queries reflect the current, updated state or acknowledge uncertainty appropriately. The module must avoid hallucinating unsupported entities or facts and must ground edges in the explicitly available and valid entities, maintaining consistent temporal and logical coherence.
\\[4pt]
\textbf{(dedupe\_nodes, appended)} If the module receives extracted entities and existing knowledge base entities, then it should resolve duplicates strictly based on identity and completeness without merging distinct entities or omitting valid ones. It should preserve distinct technical entities especially configuration parameters and versions that are critical for correctness and traceability in knowledge representation, thereby preventing loss of information critical to later query answering.

If the module receives newly extracted entities and updated knowledge base entities reflecting changed or deleted facts, then it should carefully resolve duplicates only for valid, consistent entities, respecting conditional dependencies and removing outdated facts. The module should strictly avoid merging distinct entities or preserving obsolete information, particularly when explicit deletions or updates are available, ensuring an accurate entity inventory aligned with the latest user context.
}
\end{promptbox}
\caption{Graphiti: SIMBA-appended advice blocks for the three optimized prompts.}
\label{fig:prompt-simba-graphiti}
\end{figure}

\paragraph{Karpathy Wiki.}
SIMBA appended one advice paragraph to each of the three Karpathy Wiki prompts (\texttt{flush}, \texttt{compile}, \texttt{query}). The appended texts are shown below.

\begin{figure}[H]
\begin{promptbox}
{\small
\textbf{(flush, appended)} If the module receives raw conversation data containing key project facts, then it should focus on extracting and summarizing actionable, non-trivial information precisely and completely. It should emphasize capturing explicit facts about configurations, deployments, and changes, and flagging any deletions or updates to ensure downstream modules have accurate logs to compile from.
\\[4pt]
\textbf{(compile, appended)} If the module receives daily logs including configuration details and knowledge updates, then it should rigorously extract all relevant factual knowledge, including explicit mappings of infrastructure dependencies and conditional changes. It should represent these with clarity in the wiki, explicitly noting dependencies and if-then scenarios, and carefully documenting deletions or overrides. Complete YAML frontmatter and linking related concepts in the knowledge base are essential for quality.
\\[4pt]
\textbf{(query, appended)} If the module receives questions or queries about configuration, deployment, or error logs, then it should perform context-aware retrieval prioritizing exact matches and dependency-aware reasoning. It should handle deletions properly, recognize conditional dependencies (e.g., if deploy target changes, then CI config changes), and be capable of precise recall for exact error logs. Summarizing complex dependencies succinctly while citing relevant articles or sources can improve answer accuracy and completeness.
}
\end{promptbox}
\caption{Karpathy Wiki: SIMBA-appended advice blocks for the three optimized prompts.}
\label{fig:prompt-simba-karpathy}
\end{figure}

Per-task stability statistics for the SIMBA baseline-vs-optimized comparison are reported in \cref{app:sd-simba}, computed in the same $N{=}5$ format as \cref{tab:sd}.

\section{Noise Robustness Details}
\label{app:noise}

We evaluate three representative systems (MD-flat, Mem0, and text-embedding-3-small) under three filler conditions: no filler (evidence sessions only), 32K tokens (default), and 128K tokens (4$\times$ default). Each condition uses the same 40 episodes (20 PL + 20 SW). Deletion, Cascade, and Absence scores use the same trivial-pass filtering as the main results table.

\begin{figure}[H]
\centering
\begin{tikzpicture}
\begin{axis}[
    name=plot1,
    ybar, bar width=13pt, width=0.95\textwidth, height=4.2cm,
    nodes near coords, every node near coord/.append style={font=\tiny},
    ylabel={Accuracy}, title={\textbf{MD-flat}},
    symbolic x coords={Tracking, Aggregation, Exact Recall, Deletion, Cascade, Absence},
    xtick=data, xticklabel style={rotate=30, anchor=east, font=\small},
    ymin=0, ymax=1.15,
    ytick={0,0.5,1.0},
    legend style={at={(0.98,0.98)}, anchor=north east, font=\small},
    enlarge x limits=0.1,
    /pgf/number format/fixed, /pgf/number format/precision=2,
]
\addplot[fill=black!15] coordinates {(Tracking,0.53) (Aggregation,0.78) (Exact Recall,0.57) (Deletion,0.47) (Cascade,0.05) (Absence,0.02)};
\addplot[fill=black!40] coordinates {(Tracking,0.75) (Aggregation,0.70) (Exact Recall,0.97) (Deletion,0.20) (Cascade,0.05) (Absence,0.02)};
\addplot[fill=black!70] coordinates {(Tracking,0.55) (Aggregation,0.50) (Exact Recall,0.85) (Deletion,0.20) (Cascade,0.05) (Absence,0.02)};
\legend{No filler, 32K, 128K}
\end{axis}
\end{tikzpicture}

\vspace{0.5cm}
\begin{tikzpicture}
\begin{axis}[
    name=plot2,
    ybar, bar width=13pt, width=0.95\textwidth, height=4.2cm,
    nodes near coords, every node near coord/.append style={font=\tiny},
    ylabel={Accuracy}, title={\textbf{Mem0}},
    symbolic x coords={Tracking, Aggregation, Exact Recall, Deletion, Cascade, Absence},
    xtick=data, xticklabel style={rotate=30, anchor=east, font=\small},
    ymin=0, ymax=1.15,
    ytick={0,0.5,1.0},
    legend style={at={(0.98,0.98)}, anchor=north east, font=\small},
    enlarge x limits=0.1,
    /pgf/number format/fixed, /pgf/number format/precision=2,
]
\addplot[fill=black!15] coordinates {(Tracking,0.28) (Aggregation,0.82) (Exact Recall,0.60) (Deletion,0.05) (Cascade,0.03) (Absence,0.02)};
\addplot[fill=black!40] coordinates {(Tracking,0.23) (Aggregation,0.60) (Exact Recall,0.68) (Deletion,0.03) (Cascade,0.02) (Absence,0.00)};
\addplot[fill=black!70] coordinates {(Tracking,0.23) (Aggregation,0.42) (Exact Recall,0.62) (Deletion,0.03) (Cascade,0.00) (Absence,0.00)};
\legend{No filler, 32K, 128K}
\end{axis}
\end{tikzpicture}

\vspace{0.5cm}
\begin{tikzpicture}
\begin{axis}[
    name=plot3,
    ybar, bar width=13pt, width=0.95\textwidth, height=4.2cm,
    nodes near coords, every node near coord/.append style={font=\tiny},
    ylabel={Accuracy}, title={\textbf{text-embedding-3-small}},
    symbolic x coords={Tracking, Aggregation, Exact Recall, Deletion, Cascade, Absence},
    xtick=data, xticklabel style={rotate=30, anchor=east, font=\small},
    ymin=0, ymax=1.15,
    ytick={0,0.5,1.0},
    legend style={at={(0.98,0.98)}, anchor=north east, font=\small},
    enlarge x limits=0.1,
    /pgf/number format/fixed, /pgf/number format/precision=2,
]
\addplot[fill=black!15] coordinates {(Tracking,0.53) (Aggregation,0.35) (Exact Recall,0.93) (Deletion,0.28) (Cascade,0.08) (Absence,0.10)};
\addplot[fill=black!40] coordinates {(Tracking,0.38) (Aggregation,0.20) (Exact Recall,0.42) (Deletion,0.17) (Cascade,0.03) (Absence,0.14)};
\addplot[fill=black!70] coordinates {(Tracking,0.25) (Aggregation,0.15) (Exact Recall,0.23) (Deletion,0.20) (Cascade,0.03) (Absence,0.12)};
\legend{No filler, 32K, 128K}
\end{axis}
\end{tikzpicture}
\caption{Per-task accuracy under three filler conditions (no filler, 32K, 128K) for each system. Deletion, Cascade, and Absence use trivial-pass filtering (same convention as the main results table).}
\label{fig:noise-detail}
\end{figure}

\textbf{General trend (32K $\to$ 128K).}
All three systems show overall degradation when filler volume increases from 32K to 128K: MD-flat drops from 0.45 to 0.36, Mem0 from 0.26 to 0.22, and text-embedding-3-small from 0.22 to 0.16. For Mem0 and text-embedding-3-small, the decline is concentrated in retrieval tasks (Aggregation, Exact Recall), where increased filler dilutes the signal-to-noise ratio in semantic search. MD-flat's decline is broader, affecting Aggregation ($-$0.20) and Tracking ($-$0.20), as the agent's monolithic memory file grows proportionally with filler volume.

\textbf{No-filler anomaly for MD-flat.}
Counter-intuitively, MD-flat's overall accuracy is \emph{lower} without filler (0.40) than with 32K filler (0.45). Inspection reveals that the retrieve step (gpt-4.1-mini) behaves differently when the memory file is short: it strips timestamps and entity labels, returning bare bullet lists (e.g., ``- User drives Zyvanta Sedan'' instead of ``[2023/03/01] User drives Zyvanta Sedan''). The answering LLM (Claude Sonnet 4), following its strict context-only instruction, then refuses to infer chronological order or entity provenance, causing failures on Tracking and Exact Recall. With 32K filler, the memory file is longer and more diverse, prompting the retrieve step to preserve metadata. This artifact is specific to the retrieve prompt's sensitivity to memory file length rather than a property of the benchmark itself.

\textbf{Cascade and Absence remain at floor.}
Across all three noise conditions and all three systems, Cascade and Absence scores remain near zero (0.00--0.14), reinforcing the main finding that under the default configuration the dependency-reasoning gap is not driven by retrieval noise.

\textbf{Trivial-pass rates rise with noise (justifies the filter).}
The trivial-pass filter discards Absence/Deletion/Cascade ``passes'' that came from a system that never knew the original value (i.e., before-state-check failed). \Cref{tab:trivial-pass} shows the trivial-pass count per (system, condition, task). For raw retrieval (text-embedding-3-small), the Absence trivial-pass rate rises sharply with noise (0.06 $\to$ 0.24 $\to$ 0.51): under heavy filler the top-$k$ retrieved chunks are dominated by irrelevant content, the answering LLM defaults to ``I don't know'', and ``I don't know'' happens to match the Absence gold of ``uncertain.'' Without the trivial-pass filter, this artifact would inflate text-embedding-3-small's Absence score to 0.16, 0.39, and 0.63 across the three conditions; with the filter, the score stays at 0.10--0.14, matching the main finding that the dependency gap is not closed by retrieval failure.

\begin{table}[h]
\centering
\caption{Trivial-pass counts (and rates) per condition for the three noise-evaluated systems on the 40-episode subset. A trivial pass is a question where the system answered correctly post-change but failed the pre-change state-check, indicating it never knew the original value. The trivial-pass filter excludes these from the numerator of Cascade, Absence, and Deletion accuracy.}
\label{tab:trivial-pass}
\footnotesize
\renewcommand{\arraystretch}{0.85}
\begin{tabular*}{\columnwidth}{@{\extracolsep{\fill}}ll*{3}{c}@{}}
\toprule
\textbf{System} & \textbf{Condition} & \textbf{Cascade} & \textbf{Absence} & \textbf{Deletion} \\
\midrule
MD-flat                & no filler & 0/64 (0.00)  & 0/49 (0.00)   & 2/40 (0.05)  \\
                       & 32K       & 2/64 (0.03)  & 1/49 (0.02)   & 4/40 (0.10) \\
                       & 128K      & 1/64 (0.02)  & 6/49 (0.12)  & 6/40 (0.15) \\
\midrule
Mem0                   & no filler & 0/64 (0.00)  & 0/49 (0.00)   & 0/40 (0.00)  \\
                       & 32K       & 2/64 (0.03)  & 0/49 (0.00)   & 1/40 (0.02)  \\
                       & 128K      & 0/64 (0.00)  & 0/49 (0.00)   & 0/40 (0.00)  \\
\midrule
text-embedding-3-small & no filler & 3/64 (0.05)  & 3/49 (0.06)   & 6/40 (0.15) \\
                       & 32K       & 1/64 (0.02)  & 12/49 (0.24) & 14/40 (0.35) \\
                       & 128K      & 3/64 (0.05)  & 25/49 (0.51) & 19/40 (0.48) \\
\bottomrule
\end{tabular*}
\end{table}

\section{Repeated-Run Stability}
\label{app:sd}

Hosted-API LLM calls are not perfectly deterministic even at temperature$=$0 (floating-point batch-dependence, MoE routing, heterogeneous shards). We rerun two configurations under $N{=}5$ identical trials each on a 10-episode subset (5 PL + 5 SW) to (i) confirm that the single-run results are not sampling artifacts and (ii) calibrate the noise floor for the SIMBA prompt-optimization comparison in \cref{app:ablation}. Raw retrieval (BM25, text-embedding-3-small) is excluded throughout because its retrieval step is deterministic; we do not re-sample its answering and judging calls. Trivial-pass filtering applies to Cascade, Absence, and Deletion as in the main results.

\subsection{Ablation baseline}\label{app:sd-main}

We rerun ingestion, retrieval, and answering on the four systems with non-deterministic LLM components (Mem0, Graphiti, MD-flat, Karpathy Wiki) under the ablation configuration of \cref{sec:no-closure} (gpt-4.1-mini internal LLM, Claude Sonnet 4 answering LLM); $N{=}5$ trials, including the original run on the same 10 episodes plus four additional sampling trials.

\begin{table}[h]
\centering
\caption{Per-task accuracy, mean $\pm$ SD across $N{=}5$ trials on the 10-episode subset, ablation configuration.}
\label{tab:sd}
\footnotesize
\renewcommand{\arraystretch}{0.85}
\setlength{\tabcolsep}{3pt}
\begin{tabular*}{\columnwidth}{@{\extracolsep{\fill}}l*{7}{c}@{}}
\toprule
\textbf{System} & \textbf{ER} & \textbf{Agg} & \textbf{Tr} & \textbf{Del} & \textbf{Cas} & \textbf{Abs} & \textbf{Overall} \\
\midrule
Mem0          & 0.64 $\pm$ 0.05 & 0.78 $\pm$ 0.08 & 0.26 $\pm$ 0.09 & 0.02 $\pm$ 0.04 & 0.05 $\pm$ 0.03 & 0.01 $\pm$ 0.03 & 0.29 $\pm$ 0.02 \\
Graphiti      & 0.04 $\pm$ 0.05  & 0.14 $\pm$ 0.05 & 0.02 $\pm$ 0.04  & 0.04 $\pm$ 0.09 & 0.01 $\pm$ 0.02 & 0.01 $\pm$ 0.03 & 0.04 $\pm$ 0.02 \\
MD-flat       & 0.88 $\pm$ 0.08 & 0.64 $\pm$ 0.05 & 0.82 $\pm$ 0.04 & 0.08 $\pm$ 0.04 & 0.11 $\pm$ 0.03 & 0.00 $\pm$ 0.00 & 0.42 $\pm$ 0.01 \\
Karpathy Wiki & 0.10 $\pm$ 0.07 & 0.18 $\pm$ 0.08 & 0.38 $\pm$ 0.15 & 0.02 $\pm$ 0.04 & 0.01 $\pm$ 0.03 & 0.03 $\pm$ 0.04 & 0.12 $\pm$ 0.01 \\
\bottomrule
\end{tabular*}
\end{table}

The dependency gap is robust to sampling: Cascade SD $\leq$ 0.03 and Absence SD $\leq$ 0.04 across every system. Per-system Overall accuracy moves by at most 0.02 (SD), preserving the system ranking observed in the main table. Karpathy Wiki shows the largest task-level variance (Tracking SD 0.15), driven by its agentic query loop with multiple internal LLM calls per question; the other systems retrieve and answer in a fixed number of LLM calls and show smaller variance.

\subsection{SIMBA prompt-optimization (MD-flat)}\label{app:sd-simba}

For the prompt-optimization experiment described in \cref{app:ablation}, we re-evaluate the baseline and SIMBA-optimized configurations on the same 10-episode SIMBA test set under $N{=}5$ trials each. Multi-seed stability is reported only for MD-flat. It shows the largest baseline-vs-optimized change ($+0.25$ Overall), so noise-floor calibration is most informative there. For Mem0, Graphiti, and Karpathy Wiki, the single-seed baseline-vs-optimized change is $\leq 0.02$ Overall, so the qualitative finding is robust to within-system noise.

\begin{table}[h]
\centering
\caption{Per-task accuracy, mean $\pm$ SD across $N{=}5$ trials on the SIMBA 10-episode test set (MD-flat); same format as \cref{tab:sd}.}
\label{tab:simba-stability}
\footnotesize
\renewcommand{\arraystretch}{0.85}
\setlength{\tabcolsep}{3pt}
\begin{tabular*}{\columnwidth}{@{\extracolsep{\fill}}l*{7}{c}@{}}
\toprule
\textbf{Config.} & \textbf{ER} & \textbf{Agg} & \textbf{Tr} & \textbf{Del} & \textbf{Cas} & \textbf{Abs} & \textbf{Overall} \\
\midrule
Baseline  & 0.90 $\pm$ 0.07 & 0.30 $\pm$ 0.12 & 0.30 $\pm$ 0.10 & 0.02 $\pm$ 0.04 & 0.07 $\pm$ 0.03 & 0.03 $\pm$ 0.05 & 0.27 $\pm$ 0.05 \\
Optimized & 1.00 $\pm$ 0.00 & 0.78 $\pm$ 0.08 & 0.92 $\pm$ 0.04 & 0.42 $\pm$ 0.08 & 0.02 $\pm$ 0.03 & 0.00 $\pm$ 0.00 & 0.52 $\pm$ 0.02 \\
\bottomrule
\end{tabular*}
\end{table}

A clear dichotomy emerges. Tasks driven by \emph{what} the memory file contains (Tracking, Aggregation, Exact Recall, Deletion) show large gains far exceeding the noise floor: Tracking $0.30 \to 0.92$ ($+0.62$ vs SD $\leq$ 0.10), Aggregation $0.30 \to 0.78$ ($+0.48$ vs SD $\leq$ 0.12), Deletion $0.02 \to 0.42$ ($+0.40$ vs SD $\leq$ 0.08). In contrast, Cascade and Absence differences ($-0.05$ and $-0.03$) sit within the SD of either configuration.

\section{Answering LLM Swap: Per-System Breakdown}
\label{app:llm-ablation}

\Cref{tab:answer-llm-detail} reports the full per-task accuracy for the answering-LLM swap summarized in \cref{tab:answer-llm-summary}.

\begin{table*}[h]
\centering
\caption{Answer LLM ablation, per-system: replacing the answering LLM (gpt-4.1-mini $\to$ Claude Sonnet 4) on all six main-table systems and 100 episodes; internal LLM held at gpt-4.1-mini. Trivial-pass filtering applied to Cascade, Absence, Deletion. \textbf{Bold} marks the best per task column.}
\label{tab:answer-llm-detail}
\footnotesize
\renewcommand{\arraystretch}{0.85}
\setlength{\tabcolsep}{4pt}
\begin{tabular*}{\textwidth}{@{\extracolsep{\fill}}ll*{7}{c}@{}}
\toprule
\textbf{System} & \textbf{Answering LLM} & \textbf{ER} & \textbf{Agg} & \textbf{Tr} & \textbf{Del} & \textbf{Cas} & \textbf{Abs} & \textbf{Overall} \\
\midrule
BM25                   & gpt-4.1-mini & \textbf{1.00} & 0.05 & 0.16 & \textbf{0.27} & 0.02 & 0.00 & 0.25 \\
                       & Sonnet 4     & 0.70 & 0.09 & 0.11 & 0.19 & 0.01 & 0.12 & 0.20 \\
\midrule
text-embedding-3-small & gpt-4.1-mini & 0.96 & 0.33 & 0.46 & 0.17 & 0.04 & 0.00 & 0.33 \\
                       & Sonnet 4     & 0.43 & 0.27 & 0.36 & 0.20 & 0.03 & \textbf{0.16} & 0.24 \\
\midrule
Mem0                   & gpt-4.1-mini & 0.67 & 0.35 & 0.43 & 0.21 & 0.03 & 0.00 & 0.28 \\
                       & Sonnet 4     & 0.67 & 0.65 & 0.16 & 0.03 & 0.01 & 0.00 & 0.25 \\
\midrule
Graphiti               & gpt-4.1-mini & 0.03 & 0.01 & 0.04 & 0.09 & 0.02 & 0.01 & 0.03 \\
                       & Sonnet 4     & 0.03 & 0.01 & 0.00 & 0.09 & 0.04 & 0.00 & 0.03 \\
\midrule
MD-flat                & gpt-4.1-mini & 0.94 & 0.45 & 0.77 & 0.25 & \textbf{0.05} & 0.05 & 0.42 \\
                       & Sonnet 4     & 0.90 & \textbf{0.69} & \textbf{0.80} & 0.24 & \textbf{0.05} & 0.05 & \textbf{0.46} \\
\midrule
Karpathy Wiki          & gpt-4.1-mini & 0.11 & 0.18 & 0.27 & 0.03 & 0.01 & 0.02 & 0.10 \\
                       & Sonnet 4     & 0.11 & 0.18 & 0.27 & 0.03 & 0.01 & 0.02 & 0.10 \\
\bottomrule
\end{tabular*}
\end{table*}

{
\section{Per-stage Cascade Traces for the Remaining Systems}
\label{app:cascade-snapshots-extra}

The per-stage trace for Mem0, MD-flat (gpt-4.1-mini), BM25, and text-embedding-3-small on the same Cascade episode (sw\_033) used in \cref{fig:memory-snapshots}. All four systems encode the rule and the pre-change value, but each fails at a different downstream stage; the answering LLM in every case ends up reporting the pre-change value (Hyunwoo Nam).

\begin{figure}[t]
\centering
{\centering\footnotesize
\colorbox{precolor}{\rule{0pt}{0.7em}\hspace{1em}}~pre-change value\hspace{1em}%
\colorbox{rulecolor}{\rule{0pt}{0.7em}\hspace{1em}}~conditional rule\hspace{1em}%
\colorbox{chgcolor}{\rule{0pt}{0.7em}\hspace{1em}}~change event\hspace{1em}%
\colorbox{goldcolor}{\rule{0pt}{0.7em}\hspace{1em}}~answer\par}\vspace{6pt}

\begin{episodebox}
\footnotesize
\textbf{Episode sw\_033 (Software Project, Cascade).}
\par\setlength{\leftskip}{1em}\setlength{\parskip}{0pt}
Sess.\,11: \emph{``Weekly report recipient is \hPre{Hyunwoo Nam} (assigned by team lead Seokjin Kang). \hRule{If the team lead changes, the recipient will be James Lee.}''}\par
Sess.\,14: \emph{``\hChg{Jihoon Ryu is our new team lead.}''}\par
\vspace{4pt}
Question: \emph{``Who receives the weekly report?''}\quad Answer: \hGold{James Lee}.\quad \textbf{All four systems answer Hyunwoo Nam.}\par
\end{episodebox}\vspace{3pt}

\begin{minipage}[t]{0.498\textwidth}
\begin{snapbox}[equal height group=fig30eq]{BM25 -- Failure}
\footnotesize
\textbf{1.\,\stagepass\,Encoding}: all facts encoded\\[2pt]
{\tiny\hangindent=1em\hangafter=1
\hPre{Sess.\,11 (03/13): ``weekly report recipient is Hyunwoo Nam\ldots''}\\
\hRule{``\ldots if team lead changes, recipient will be James Lee''}\par}\vspace{6pt}
\textbf{2.\,\stagepass\,Maintenance}: all facts retained\\[2pt]
{\tiny\hangindent=1em\hangafter=1
\hChg{Sess.\,14 (03/17): ``Jihoon Ryu is our new team lead''}\par}\vspace{6pt}
\textbf{3.\,\stagefail\,Retrieval}: lexical top-$k$ misses the 03/17 session\\[2pt]
{\tiny\hangindent=1em\hangafter=1
top-1: \hPre{Sess.\,11 (recipient + rule)}\\
\mbox{[unrelated filler]}\par}
\end{snapbox}
\end{minipage}\hfill
\begin{minipage}[t]{0.498\textwidth}
\begin{snapbox}[equal height group=fig30eq]{MD-flat (gpt-4.1-mini) -- Failure}
\footnotesize
\textbf{1.\,\stagepass\,Encoding}: all facts encoded\\[2pt]
{\tiny\hangindent=1em\hangafter=1
\texttt{[2023/03/13]} \hPre{Weekly report recipient is Hyunwoo Nam \ldots}\\
\texttt{[2023/03/13]} \hRule{If team lead changes, recipient will be James Lee}\par}\vspace{6pt}
\textbf{2.\,\stagepass\,Maintenance}: all facts retained\\[2pt]
{\tiny\hangindent=1em\hangafter=1
\texttt{[2023/03/17]} \hChg{Jihoon Ryu is the new team lead}\par}\vspace{6pt}
\textbf{3.\,\stagefail\,Retrieval}: tool-use never opens 03/17 entry\\[2pt]
{\tiny\hangindent=1em\hangafter=1
- \hPre{Weekly report recipient is Hyunwoo Nam \ldots}\\
- \hRule{If team lead changes, recipient will be James Lee}\par}
\end{snapbox}
\end{minipage}\\[1pt]

\begin{minipage}[t]{0.498\textwidth}
\begin{snapbox}[equal height group=fig30eq2]{text-embedding-3-small -- Failure}
\footnotesize
\textbf{1.\,\stagepass\,Encoding}: all facts encoded\\[2pt]
{\tiny\hangindent=1em\hangafter=1
\hPre{Sess.\,11 (03/13): ``weekly report recipient is Hyunwoo Nam\ldots''}\\
\hRule{``\ldots if team lead changes, recipient will be James Lee''}\par}\vspace{6pt}
\textbf{2.\,\stagepass\,Maintenance}: all facts retained\\[2pt]
{\tiny\hangindent=1em\hangafter=1
\hChg{Sess.\,14 (03/17): ``Jihoon Ryu is our new team lead''}\par}\vspace{6pt}
\textbf{3.\,\stagepass\,Retrieval}: change event surfaced\\[2pt]
{\tiny\hangindent=1em\hangafter=1
top-1: \hPre{Sess.\,11 (recipient + rule)}\\
top-2: \hChg{Sess.\,14 (Jihoon Ryu)}\\
\mbox{[unrelated filler]}\par}\vspace{6pt}
\textbf{4.\,\stagefail\,Answer}: LLM reports \hPre{Hyunwoo Nam}
\end{snapbox}
\end{minipage}\hfill
\begin{minipage}[t]{0.498\textwidth}
\begin{snapbox}[equal height group=fig30eq2]{Mem0 (gpt-4.1-mini) -- Failure}
\footnotesize
\textbf{1.\,\stagepass\,Encoding}: all facts encoded\\[2pt]
{\tiny\hangindent=1em\hangafter=1
User's weekly report recipient is \hPre{Hyunwoo Nam} \ldots, \hRule{with the recipient changing to James Lee if the team lead changes}\par}\vspace{6pt}
\textbf{2.\,\stagepass\,Maintenance}: all facts retained\\[2pt]
{\tiny\hangindent=1em\hangafter=1
\hChg{User's new team lead is Jihoon Ryu \ldots}\par}\vspace{6pt}
\textbf{3.\,\stagepass\,Retrieval}: change event surfaced\\[2pt]
{\tiny\hangindent=1em\hangafter=1
\texttt{[0.65]} \hPre{Hyunwoo Nam} \ldots \hRule{changing to James Lee if the team lead changes}\\
\mbox{[unrelated filler]}\\
\texttt{[0.25]} \hChg{Jihoon Ryu \ldots} (rank 19/20)\par}\vspace{6pt}
\textbf{4.\,\stagefail\,Answer}: LLM reports \hPre{Hyunwoo Nam}
\end{snapbox}
\end{minipage}

\caption{State of the remaining four memory systems on episode sw\_033, traced across encoding, maintenance, retrieval, and the answering LLM. Top row: BM25 and MD-flat (gpt-4.1-mini) drop the change event before retrieval. Bottom row: text-embedding-3-small and Mem0 surface the change event in the retrieved context, but the answering LLM still reports the pre-change value. Same color coding as \cref{fig:memory-snapshots}.}
\label{fig:memory-snapshots-extra}
\end{figure}
}

\section{Retrieval vs.\ Reasoning Bottleneck on Cascade and Absence (Top-$k$ Sweep)}
\label{app:retrieval-vs-reasoning}

For each \emph{knew\_but\_failed} case in the top-$k$ sweep (BM25 and dense, 40-episode subset, Claude Sonnet 4 answering LLM), we check whether both the dependency rule and the upstream change event were already in the retrieval context. A failure where both are present is an answering failure (both rule and change are in the retrieved context but the answering LLM still fails to propagate); a failure where one is missing is a retrieval failure.

\paragraph{Method.} We use the dataset's gold-fact metadata to identify, for each task, the specific evidence sessions carrying the rule and the change event. The rule session contains a gold-fact entry with \texttt{entity = target\_entity} and (Cascade) \texttt{is\_if\_then = True} or (Absence) \texttt{has\_dependency = True}. The change session is the canonical \texttt{evidence\_change+delete\_event} session, with a gold-fact entry where \texttt{entity = cascade\_source} and \texttt{type = root\_change}. For each rule and change fact, the \texttt{original\_seed} string appears verbatim in its session's conversation in 100\% of cases (213 rule, 100 change, 341 dependency facts in the 100-episode dataset), so checking for \texttt{original\_seed} as a substring of the retrieval context is exact. Identical breakdowns are obtained by matching the session's \texttt{[Session: <timestamp>]} header instead, confirming that no rule/change content is split off from its session header by chunking.

\paragraph{Results.} \Cref{tab:retrieval-vs-reasoning} partitions \emph{knew\_but\_failed} cases by which stage missed the propagated answer: change-event missing from retrieval, rule missing from retrieval, or both retrieved but the answering LLM fails to propagate. The rule is retrieved in 100\% of failures across all $k$, so retrieval-side losses are change-event misses; the live trade-off at higher $k$ is between change-event miss and answering failure.
\begin{itemize}[leftmargin=1.4em,itemsep=1pt,topsep=2pt]
\item Cascade: at $k{=}20$ and $k{=}40$, 55\% of failures are change-event misses and 45\% are answering failures (the rule and the change session are both retrieved yet the answering LLM still does not produce the propagated value).
\item Absence: at $k{=}20$ and $k{=}40$, 86\% and 83\% of failures respectively are answering failures (the agent commits to a definite answer when it should abstain), with the remainder being change-event misses.
\end{itemize}

\begin{table}[h]
\centering
\caption{\emph{Knew\_but\_failed} cases on the top-$k$ sweep, partitioned by failure stage (gold-fact substring match). Combined BM25 + dense on a 40-episode subset (128 Cascade and 98 Absence case-system pairs) with Sonnet 4 as the answering LLM. \emph{Cases} counts knew\_but\_failed events; the remaining failures are never\_knew (system did not encode the fact before the change). Knew\_but\_failed accounts for 17--45\% of the Cascade pool and 7--57\% of the Absence pool across the four $k$ values. Per-cell percentages are within \emph{Cases}: the Retrieval group splits change-event miss from rule miss, and Answering counts cases where both rule and change are in the retrieved context but the answering LLM still fails to propagate.}
\label{tab:retrieval-vs-reasoning}
\footnotesize
\renewcommand{\arraystretch}{0.85}
\setlength{\tabcolsep}{6pt}
\begin{tabular}{llcccc}
\toprule
& & & \multicolumn{2}{c}{\textbf{Retrieval}} & \\
\cmidrule(lr){4-5}
\textbf{Task} & \textbf{top-$k$} & \textbf{Cases} & \textbf{Change} & \textbf{Rule} & \textbf{Answering} \\
\midrule
Cascade  &  5 & 57 & 32 (56\%) & 0 (0\%) & 25 (44\%) \\
         & 10 & 34 & 10 (29\%) & 0 (0\%) & 24 (71\%) \\
         & 20 & 22 & 12 (55\%) & 0 (0\%) & 10 (45\%) \\
         & 40 & 22 & 12 (55\%) & 0 (0\%) & 10 (45\%) \\
\midrule
Absence  &  5 & 56 & 32 (57\%) & 0 (0\%) & 24 (43\%) \\
         & 10 & 16 &  7 (44\%) & 0 (0\%) &  9 (56\%) \\
         & 20 &  7 &  1 (14\%) & 0 (0\%) &  6 (86\%) \\
         & 40 & 12 &  2 (17\%) & 0 (0\%) & 10 (83\%) \\
\bottomrule
\end{tabular}
\end{table}

\section{MD-flat Internal-LLM Mechanisms}
\label{app:mdflat-internal}

This section traces the per-LLM internal-storage mechanisms that explain the Cascade and Absence outcomes of the internal-LLM ablation in \cref{tab:llm-ablation}.

\subsection{MD-flat with gpt-5: ingestion compresses memory and erases prior entries}
\label{qual:mdflat-gpt5-compression}

\Cref{tab:write-behavior} summarizes \texttt{write\_memory} behavior under the four internal LLMs on the 20-episode subset. The three stronger LLMs (gpt-5, GLM-5.1, Opus 4.7) treat Change+Delete sessions as memory-update triggers (95--100\% write rate), while gpt-4.1-mini almost always uses \texttt{append\_memory}. Comparing memory file sizes \emph{before} and \emph{after} the Change+Delete write reveals the per-LLM compression strategy: gpt-5 compresses on every write (cumulatively shrinking the pre-event memory to ${\sim}1{,}800$ chars and dropping a further $28\%$ at the change event), GLM-5.1 is effectively append-only ($+1\%$ at the event, so prior entries are preserved), and Opus 4.7 restructures while keeping explicit \texttt{Contingency:} entries ($-5\%$ at the event; mechanism in \cref{qual:mdflat-opus-contingency}). gpt-4.1-mini barely writes, so its memory just accumulates session by session.

\begin{table}[h]
\centering
\caption{\texttt{write\_memory} invocations and memory file length around the Change+Delete session on MD-flat (20 episodes, internal-LLM ablation). The Before/After columns show average memory file size just before and after the Change+Delete write; the delta exposes each LLM's compression strategy.}
\label{tab:write-behavior}
\footnotesize
\renewcommand{\arraystretch}{0.85}
\begin{tabular*}{\columnwidth}{@{\extracolsep{\fill}}lcccc@{}}
\toprule
\textbf{Internal LLM} & \textbf{Change+Delete} & \textbf{Memory before} & \textbf{Memory after} & \textbf{$\Delta$} \\
              & \textbf{write rate}    & \textbf{(chars)}        & \textbf{(chars)}       &                    \\
\midrule
gpt-4.1-mini & 2/20 (10\%)   & ${\sim}7{,}500$ & ${\sim}7{,}400$ & ${\approx}0\%$  \\
gpt-5        & 19/20 (95\%)  & ${\sim}1{,}800$ & ${\sim}1{,}300$ & $-28\%$         \\
GLM-5.1      & 19/20 (95\%)  & ${\sim}5{,}100$ & ${\sim}5{,}100$ & $+1\%$          \\
Opus 4.7     & 20/20 (100\%) & ${\sim}2{,}900$ & ${\sim}2{,}700$ & $-5\%$          \\
\bottomrule
\end{tabular*}
\end{table}

Trigger detection alone does not yield dependency reasoning. The downstream effect of gpt-5's lossy compression is visible in the Absence column of \cref{tab:llm-ablation}: gpt-5's compressed memory often contains neither the original entity value nor the dependency rule, so retrieval returns ``(no relevant facts)'' and the answering LLM defaults to ``I don't have that information,'' which coincidentally matches the Absence gold of ``uncertain.'' GLM-5.1's append-only memory preserves prior entries, so the original value remains retrievable and the system commits to it, failing Absence.

As a concrete instance, on Personal Life Episode~7 (commute Absence question), gpt-5 reads the 4{,}440-char memory file (32 entries through 2023/03/22 including the commute fact) and then issues \texttt{write\_memory} with 236 chars containing only the four new 2023/03/24 entries (move to Corenth Hills, gaming PC purchase, Dranthazon removal, housewarming party). The prior commute fact is deleted along with 31 other entries; the after-Absence retrieval returns \texttt{(no relevant facts)}, and the agent answers ``I don't have that information.''

\subsection{MD-flat with Opus 4.7: explicit contingencies and active dependency propagation}
\label{qual:mdflat-opus-contingency}

Unlike gpt-5's lossy compression or GLM-5.1's append-only behavior, Opus 4.7 reorganizes the MD-flat memory file into a structured document with topical \texttt{\#\#} sections (e.g., \emph{Travel History}, \emph{Travel Preferences}) and, critically, records each dependency rule as an explicit \texttt{Contingency:} entry rather than embedding it inside the dependent fact. On Personal Life Episode~9, for instance, the pre-change memory contains:
\begin{quote}
\small
\texttt{[2023/03/19] Contingency: if health condition changes, exercise routine becomes yoga 2x/week.}\\
\texttt{[2023/03/19] Contingency: if health condition changes, dietary restriction becomes no alcohol.}\\
\texttt{[2023/03/19] Contingency: if exercise routine changes, will work out at Crysthene Pool.}
\end{quote}

When an upstream change arrives during a later session, Opus reacts in one of two ways depending on whether a replacement rule is available, and these two strategies are exactly what closes Cascade and Absence respectively:

\textbf{Strategy A (Cascade, replacement rule exists): apply the rule, write the resolved value.} Continuing the Episode~9 trace, after the user reports that their \emph{tendinitis} has resolved, Opus writes:
\begin{quote}
\small
\texttt{[2023/03/25] Tendinitis treated and resolved; per contingencies, dietary restriction now no alcohol and exercise routine now yoga 2x/week at Crysthene Pool.}
\end{quote}
The resolved 1-hop and 2-hop values (\emph{no alcohol}, \emph{Crysthene Pool}) are now in memory; at query time the answering LLM retrieves them directly and matches the Cascade gold answers.

\textbf{Strategy B (Absence, no replacement rule): remove dependent facts and flag re-confirmation needed.} On Software Project Episode~2, the deploy target changes from \emph{Narvex Cloud} to \emph{Thandrel Infra}, but the dependent facts (logs drain, Docker image, CI config, Dockerfile, deploy command) have no propagation rules in the memory file. Opus's post-change memory contains a new section:
\begin{quote}
\small
\texttt{\#\# Deployment-derived facts (need re-confirmation since deploy target changed to Thandrel Infra)}\\
\texttt{- Previous (Narvex Cloud) derived facts removed; ask user for updated logs drain, Docker image, CI config, Dockerfile, deploy command under Thandrel Infra}
\end{quote}
At query time, the answering LLM sees no resolved value for these entities and answers ``I don't have that information,'' which matches the Absence gold of ``uncertain.''

The other internal LLMs do not implement this two-strategy split: gpt-4.1-mini and GLM-5.1 leave stale dependent facts intact (failing Cascade and Absence), and gpt-5 compresses memory aggressively at the upstream-change session (yielding accidental Absence matches but no genuine Cascade propagation; \cref{qual:mdflat-gpt5-compression}). Opus's combination of explicit contingencies plus active propagation explains the Cascade jump from 0.00 to 0.32 and the Absence jump from 0.08 to 0.60 in the MD-flat row of \cref{tab:llm-ablation}. The same LLM produces no comparable gain on Mem0 (atomic facts decompose the contingency entries) or Graphiti (extraction discards the contingency wording), confirming that the file-based architecture is what allows Opus's reasoning to land in the memory representation.

The cost of this behavior is twofold. First, Opus's hierarchical reorganization paraphrases content during ingest, dropping Exact Recall from 0.95 to 0.60 and Tracking from 0.70 to 0.20 on the same 20-episode subset. Second, the per-episode dollar cost is roughly 23 times the gpt-4.1-mini baseline (\cref{tab:llm-cost}); both deltas are reported in \cref{tab:llm-ablation}.

\section{In-Context Ceiling: Gold Facts Only}
\label{app:in-context-ceiling}

The gold-facts in-context ceiling feeds only the task-relevant gold facts (no filler) directly to the answering LLM. It confirms that the tasks are solvable in principle when retrieval is perfect.

\begin{table}[h]
\centering
\footnotesize
\renewcommand{\arraystretch}{0.85}
\caption{Per-task accuracy of the gold-facts in-context ceiling (no memory system, no filler) across four answer LLMs. Best per task in \textbf{bold}.}
\label{tab:in-context-ceiling}
\begin{tabular*}{\columnwidth}{@{\extracolsep{\fill}}l*{7}{c}@{}}
\toprule
\textbf{Answering LLM} & \textbf{ER} & \textbf{Agg} & \textbf{Tr} & \textbf{Del} & \textbf{Cas} & \textbf{Abs} & \textbf{Overall} \\
\midrule
Opus 4.7       & \textbf{1.00} & 0.85          & 0.96          & \textbf{0.98} & \textbf{0.93} & 0.73          & \textbf{0.91} \\
Sonnet 4.6     & \textbf{1.00} & 0.87          & \textbf{0.98} & 0.70          & 0.84          & 0.62          & 0.84 \\
Sonnet 4       & \textbf{1.00} & \textbf{0.95} & 0.97          & 0.87          & 0.60          & \textbf{0.80} & 0.87 \\
gpt-4.1-mini   & \textbf{1.00} & 0.47          & \textbf{0.98} & 0.62          & 0.74          & 0.37          & 0.70 \\
\bottomrule
\end{tabular*}
\end{table}

\begin{table}[h]
\centering
\footnotesize
\renewcommand{\arraystretch}{0.85}
\caption{Ceiling-normalized accuracy: each task accuracy in \cref{tab:main_results} divided by the gold-facts in-context ceiling of gpt-4.1-mini on the same task (ER 1.00, Agg 0.47, Tr 0.98, Del 0.62, Cas 0.74, Abs 0.37; \cref{tab:in-context-ceiling}). Overall is the unweighted mean of the six normalized values. The ranking matches \cref{tab:main_results}.}
\label{tab:ceiling-normalized}
\begin{tabular*}{\columnwidth}{@{\extracolsep{\fill}}l*{6}{c}c@{}}
\toprule
\textbf{System} & \textbf{ER} & \textbf{Agg} & \textbf{Tr} & \textbf{Del} & \textbf{Cas} & \textbf{Abs} & \textbf{Overall} \\
\midrule
gpt-4.1-mini in-context & 1.00 & 0.57 & 0.70 & 0.73 & 0.04 & 0.09 & 0.52 \\
\midrule
BM25                    & 1.00 & 0.11 & 0.16 & 0.44 & 0.03 & 0.00 & 0.29 \\
text-embedding-3-small  & 0.96 & 0.70 & 0.47 & 0.27 & 0.05 & 0.00 & 0.41 \\
Mem0                    & 0.67 & 0.74 & 0.44 & 0.34 & 0.04 & 0.00 & 0.37 \\
Graphiti                & 0.03 & 0.02 & 0.04 & 0.15 & 0.03 & 0.02 & 0.05 \\
MD-flat                 & 0.94 & 0.96 & 0.79 & 0.40 & 0.07 & 0.16 & 0.55 \\
Karpathy Wiki           & 0.11 & 0.38 & 0.28 & 0.05 & 0.01 & 0.07 & 0.15 \\
\bottomrule
\end{tabular*}
\end{table}

\section{Additional Experiments}
\label{app:additional}

This appendix reports four follow-up analyses of the paper's central finding that memory systems fail Cascade and Absence. We test (i) sensitivity to the hand-crafted dependency graphs, (ii) the effect of stating dependency rules implicitly, (iii) whether a task-specific answer prompt recovers the failures, and (iv) seed stability of the internal-LLM ablation. The failure persists under (i) and (ii), the prompt in (iii) recovers only the answering-stage failures, and (iv) preserves the ordering of \cref{tab:llm-ablation}.

\subsection{Sensitivity to the Dependency-Graph Structure}
\label{app:dag-sensitivity}

This section details the structural slices summarized in \cref{sec:main-results}. Our episodes are generated from two hand-crafted DAGs, which raises the question of
whether the failure is specific to those graphs. We decompose the main-table results
(100 episodes, gpt-4.1-mini answering) along four structural dimensions: domain, hop
length, dependency pattern, and root entity. The failure appears in every slice, and
the system ranking is largely preserved, with MD-flat first and Graphiti last in
every domain and root slice. Notably, the failure does not weaken at short chains:
it already saturates at 1-hop propagation, the simplest possible case, so it is not
a consequence of long-chain difficulty.

\begin{table}[h]
\centering
\footnotesize
\renewcommand{\arraystretch}{0.85}
\caption{Cascade/Absence accuracy sliced by domain and by hop length, with the
six-task system ranking per domain. The failure is present in every slice.}
\label{tab:dag-domain-hop}
\begin{tabular*}{\columnwidth}{@{\extracolsep{\fill}}llcc l@{}}
\toprule
\textbf{Slice} & & \textbf{Cas} & \textbf{Abs} & \textbf{System ranking (best $\rightarrow$ worst)} \\
\midrule
Domain & Overall & 0.03 & 0.02 & MD-flat $>$ dense $>$ Mem0 $>$ BM25 $>$ Karpathy $>$ Graphiti \\
       & PL      & 0.04 & 0.01 & MD-flat $>$ Mem0 $>$ dense $>$ BM25 $>$ Karpathy $>$ Graphiti \\
       & SW      & 0.02 & 0.02 & MD-flat $>$ dense $>$ BM25 $>$ Mem0 $>$ Karpathy $>$ Graphiti \\
\midrule
Hop    & 1-hop   & 0.03 & 0.01 & --- \\
       & 2-hop   & 0.03 & 0.02 & --- \\
\bottomrule
\end{tabular*}
\end{table}

\begin{table}[h]
\centering
\footnotesize
\setlength{\tabcolsep}{4pt}
\renewcommand{\arraystretch}{0.85}
\caption{Cascade/Absence accuracy by root entity and by dependency pattern, both
restricted to slices with at least 10 questions. The failure holds in every slice.}
\label{tab:dag-root-pattern}
\begin{tabular}{@{}l@{\hspace{3em}}r@{\hspace{2em}}r@{}}
\toprule
& \textbf{Cas} & \textbf{Abs} \\
\midrule
\multicolumn{3}{l}{\textit{By root entity --- Personal Life}} \\
~~employer             & 0.10 & 0.00 \\
~~residence\_location  & 0.06 & 0.00 \\
~~relationship\_status & 0.03 & 0.01 \\
~~school               & 0.00 & 0.00 \\
~~health\_condition    & 0.00 & 0.03 \\
\midrule
\multicolumn{3}{l}{\textit{By root entity --- Software Project}} \\
~~team\_lead           & 0.06 & 0.00 \\
~~database             & 0.04 & 0.02 \\
~~auth\_provider       & 0.03 & 0.02 \\
~~deploy\_target       & 0.01 & 0.04 \\
~~framework            & 0.00 & 0.00 \\
\midrule
\multicolumn{3}{l}{\textit{By dependency pattern}} \\
~~distance             & 0.15 & 0.00 \\
~~company\_policy      & 0.08 & 0.00 \\
~~team\_assignment     & 0.06 & 0.00 \\
~~data\_layer          & 0.04 & 0.02 \\
~~life\_event          & 0.03 & 0.01 \\
~~derived\_config      & 0.03 & 0.02 \\
~~auth\_coupling       & 0.02 & 0.02 \\
~~tech\_compatibility  & 0.00 & 0.00 \\
~~medical\_causation   & 0.00 & 0.03 \\
~~infra\_coupling      & 0.00 & 0.04 \\
\bottomrule
\end{tabular}
\end{table}

\subsection{Implicit Dependency Rules}
\label{app:implicit}

This section details the implicit-phrasing analysis summarized in \cref{sec:main-results}. We re-generated a
20-episode subset in which only the current dependency is stated and the
forward-looking cue is removed, forcing the system to infer the trigger.

At the system level the results are essentially unchanged, because the memory systems
already sit near zero on Cascade and Absence under the explicit framing, leaving no
room to drop further. The effect of implicit phrasing is visible only at the
in-context ceiling, where the answering LLM receives the gold facts directly: Absence
falls sharply while Cascade is largely unaffected. Explicit framing therefore does
help, but only at a ceiling the memory systems never reach, so it does not inflate
the system-level results reported in the main table.

\begin{table}[h]
\centering
\footnotesize
\setlength{\tabcolsep}{4pt}
\renewcommand{\arraystretch}{0.85}
\caption{Explicit vs.\ implicit dependency phrasing on a 20-episode subset. Memory
systems answer with gpt-4.1-mini; the in-context rows give the answering LLM the
gold facts directly.}
\label{tab:implicit}
\begin{tabular*}{\columnwidth}{@{\extracolsep{\fill}}l*{4}{c}@{}}
\toprule
& \multicolumn{2}{c}{\textbf{Cas}} & \multicolumn{2}{c}{\textbf{Abs}} \\
\cline{2-3}\cline{4-5}
\addlinespace[3pt]
& \textbf{exp} & \textbf{imp} & \textbf{exp} & \textbf{imp} \\
\midrule
\multicolumn{5}{l}{\textit{Raw retrieval}} \\
~~BM25                   & 0.00 & 0.00 & 0.00 & 0.00 \\
~~text-embedding-3-small & 0.00 & 0.00 & 0.00 & 0.04 \\
\midrule
\multicolumn{5}{l}{\textit{LLM-processed memory}} \\
~~Mem0                   & 0.00 & 0.00 & 0.00 & 0.00 \\
~~Graphiti               & 0.03 & 0.00 & 0.00 & 0.00 \\
\midrule
\multicolumn{5}{l}{\textit{File-based agents}} \\
~~MD-flat                & 0.00 & 0.00 & 0.08 & 0.08 \\
~~Karpathy Wiki          & 0.00 & 0.03 & 0.04 & 0.00 \\
\midrule
\multicolumn{5}{l}{\textit{In-context ceiling (gold facts)}} \\
~~gpt-4.1-mini           & 0.52 & 0.52 & 0.37 & 0.04 \\
~~Sonnet 4.6             & 0.74 & 0.68 & 0.64 & 0.20 \\
\bottomrule
\end{tabular*}
\end{table}

\subsection{Task-Specific Answer Prompt}
\label{app:multistep}

This analysis accompanies \cref{sec:no-closure}. It uses the first ten episodes of each domain (20 in total) and gpt-4.1-mini as the answering LLM. For the five systems with a separate answering call, we reuse each system's stored retrieval context from the main run and regenerate only the answer, so retrieval is unchanged. The task-specific prompt (\cref{fig:prompt-taskspecific}) keeps the grounding instruction of the single-step prompt, removes its instruction to answer with the value alone, and adds a reasoning step followed by a final-answer line, which is the only part we score. Karpathy Wiki answers inside its own query agent, so we add the same reasoning step to its query prompt and re-run the episodes end to end.

\Cref{tab:multistep} reports all six tasks. Cascade and Absence rise for text-embedding-3-small and Mem0, the two systems whose failures \cref{sec:where-loss} traces to the answering stage, while MD-flat and Karpathy Wiki stay at or near zero and BM25 and Graphiti gain at most 0.16. Static accuracy falls for the raw retrievers, most sharply on Exact Recall (BM25 $1.00 \rightarrow 0.55$, text-embedding-3-small $0.95 \rightarrow 0.55$). The reasoning text raises output to 196{,}051 tokens over 1{,}190 questions, at most about \$0.003 per episode for gpt-4.1-mini.

\begin{figure}[h]
\begin{promptbox}
{\small
Answer the user's question based ONLY on the context provided below. If the information is not in the context, say you don't have that information.

Before answering, reason step by step: first identify any dependency rules in the context, then check whether the source entity has changed, then state the current value.

After reasoning, give the final answer on its own last line, prefixed with ``Answer:'', containing ONLY the value.

Context: \{context\}

Question: \{question\}
}
\end{promptbox}
\caption{Task-specific answer prompt used in \cref{app:multistep}. The first two lines are identical to the single-step prompt; its instruction to answer with the value only is replaced by the reasoning and final-answer lines.}
\label{fig:prompt-taskspecific}
\end{figure}

\begin{table}[h]
\centering
\scriptsize
\setlength{\tabcolsep}{3pt}
\renewcommand{\arraystretch}{0.85}
\caption{Single-step $\rightarrow$ task-specific answer prompt (20 episodes, gpt-4.1-mini answering, scoring as in \cref{tab:main_results}).}
\label{tab:multistep}
\begin{tabular*}{\columnwidth}{@{\extracolsep{\fill}}l*{6}{c}@{}}
\toprule
\textbf{System} & \textbf{ER} & \textbf{Agg} & \textbf{Tr} & \textbf{Del} & \textbf{Cas} & \textbf{Abs} \\
\midrule
BM25                   & 1.00 $\rightarrow$ 0.55 & 0.15 $\rightarrow$ 0.15 & 0.25 $\rightarrow$ 0.25 & 0.45 $\rightarrow$ 0.65 & 0.00 $\rightarrow$ 0.16 & 0.00 $\rightarrow$ 0.12 \\
text-embedding-3-small & 0.95 $\rightarrow$ 0.55 & 0.35 $\rightarrow$ 0.45 & 0.45 $\rightarrow$ 0.55 & 0.15 $\rightarrow$ 0.55 & 0.03 $\rightarrow$ 0.34 & 0.00 $\rightarrow$ 0.42 \\
Mem0                   & 0.65 $\rightarrow$ 0.65 & 0.45 $\rightarrow$ 0.50 & 0.45 $\rightarrow$ 0.60 & 0.15 $\rightarrow$ 0.45 & 0.00 $\rightarrow$ 0.12 & 0.00 $\rightarrow$ 0.38 \\
Graphiti               & 0.05 $\rightarrow$ 0.05 & 0.05 $\rightarrow$ 0.05 & 0.05 $\rightarrow$ 0.10 & 0.10 $\rightarrow$ 0.15 & 0.03 $\rightarrow$ 0.09 & 0.00 $\rightarrow$ 0.12 \\
MD-flat                & 1.00 $\rightarrow$ 1.00 & 0.55 $\rightarrow$ 0.60 & 0.85 $\rightarrow$ 0.80 & 0.15 $\rightarrow$ 0.15 & 0.00 $\rightarrow$ 0.00 & 0.00 $\rightarrow$ 0.00 \\
Karpathy Wiki          & 0.05 $\rightarrow$ 0.00 & 0.10 $\rightarrow$ 0.20 & 0.15 $\rightarrow$ 0.20 & 0.05 $\rightarrow$ 0.15 & 0.00 $\rightarrow$ 0.00 & 0.04 $\rightarrow$ 0.04 \\
\bottomrule
\end{tabular*}
\end{table}

\subsection{Seed Stability of the Internal-LLM Ablation}
\label{app:internal-llm-seeds}

\Cref{tab:llm-ablation} is a single run on a 20-episode subset. We re-ran it with three seeds on the same episodes under two constraints. The original answering LLM (Claude Sonnet 4) has since been deprecated, so all re-run cells use its successor, Claude Sonnet 4.6. We also excluded the two most expensive cells, Mem0 $\times$ Opus 4.7 and Graphiti $\times$ Opus 4.7, whose three runs we estimate at \$122 and \$743 from their per-episode costs in \cref{tab:llm-cost}; both stay near zero on Cascade and Absence in \cref{tab:llm-ablation}. The gpt-4.1-mini baseline was not re-run.

Because the answering LLM differs, absolute values are not directly comparable to \cref{tab:llm-ablation}; for example, MD-flat $\times$ Opus 4.7 Absence moves from 0.60 to $0.43 \pm 0.02$. Across-seed standard deviations are at most 0.06 in 38 of 49 cells. The ordering is unchanged: MD-flat $\times$ Opus 4.7 remains the highest-scoring pairing on both Cascade and Absence, while every re-run Mem0 and Graphiti pairing stays at or near zero on both.

\begin{table}[h]
\centering
\scriptsize
\setlength{\tabcolsep}{3pt}
\renewcommand{\arraystretch}{0.85}
\caption{Internal-LLM ablation re-run with three seeds (same 20 episodes as \cref{tab:llm-ablation}, mean $\pm$ sd, Claude Sonnet 4.6 answering). Cost is the measured total across the three runs, priced as in \cref{tab:llm-cost}.}
\label{tab:internal-llm-seeds}
\begin{tabular*}{\columnwidth}{@{\extracolsep{\fill}}ll*{7}{c}r@{}}
\toprule
\textbf{System} & \textbf{Internal LLM} & \textbf{ER} & \textbf{Agg} & \textbf{Tr} & \textbf{Del} & \textbf{Cas} & \textbf{Abs} & \textbf{Overall} & \textbf{Cost} \\
\midrule
Mem0     & gpt-5    & 0.78\pmsd{0.31} & 0.43\pmsd{0.13} & 0.32\pmsd{0.15} & 0.07\pmsd{0.05} & 0.03\pmsd{0.03} & 0.00\pmsd{0.00} & 0.27\pmsd{0.11} & \$19 \\
         & GLM-5.1  & 0.78\pmsd{0.08} & 0.58\pmsd{0.05} & 0.27\pmsd{0.08} & 0.03\pmsd{0.02} & 0.04\pmsd{0.03} & 0.00\pmsd{0.00} & 0.28\pmsd{0.01} & \$10 \\
\midrule
Graphiti & gpt-5    & 0.02\pmsd{0.02} & 0.00\pmsd{0.00} & 0.00\pmsd{0.00} & 0.02\pmsd{0.02} & 0.01\pmsd{0.02} & 0.00\pmsd{0.00} & 0.01\pmsd{0.01} & \$96 \\
         & GLM-5.1  & 0.05\pmsd{0.04} & 0.03\pmsd{0.02} & 0.02\pmsd{0.02} & 0.05\pmsd{0.07} & 0.03\pmsd{0.03} & 0.03\pmsd{0.04} & 0.03\pmsd{0.01} & \$44 \\
\midrule
MD-flat  & gpt-5    & 0.28\pmsd{0.06} & 0.25\pmsd{0.04} & 0.15\pmsd{0.04} & 0.43\pmsd{0.12} & 0.08\pmsd{0.04} & 0.24\pmsd{0.09} & 0.24\pmsd{0.01} & \$16 \\
         & GLM-5.1  & 0.83\pmsd{0.08} & 0.45\pmsd{0.04} & 0.22\pmsd{0.02} & 0.60\pmsd{0.04} & 0.20\pmsd{0.04} & 0.25\pmsd{0.04} & 0.43\pmsd{0.02} & \$7 \\
         & Opus 4.7 & 0.67\pmsd{0.05} & \textbf{0.68}\pmsd{0.06} & 0.12\pmsd{0.06} & \textbf{0.80}\pmsd{0.04} & \textbf{0.38}\pmsd{0.11} & \textbf{0.43}\pmsd{0.02} & \textbf{0.51}\pmsd{0.00} & \$92 \\
\bottomrule
\end{tabular*}
\end{table}


\medskip

{
\small
\begin{thebibliography}{24}
\providecommand{\url}[1]{\texttt{#1}}
\expandafter\ifx\csname urlstyle\endcsname\relax
  \providecommand{\doi}[1]{doi: #1}\else
  \providecommand{\doi}{doi: \begingroup \urlstyle{rm}\Url}\fi

\bibitem{chhikara2025mem0}
Chhikara, P., Khant, D., Aryan, S., Singh, T., and Yadav, D.
\newblock {Mem0}: Building production-ready {AI} agents with scalable long-term memory.
\newblock In \emph{Proceedings of the European Conference on Artificial Intelligence (ECAI)}, 2025.

\bibitem{cohen1960kappa}
Cohen, J.
\newblock A coefficient of agreement for nominal scales.
\newblock \emph{Educational and Psychological Measurement}, 20(1):37--46, 1960.

\bibitem{edge2024graphrag}
Edge, D., Trinh, H., Cheng, N., Bradley, J., Chao, A., Mody, A., Truitt, S., Metropolitansky, D., Ness, R.~O., and Larson, J.
\newblock From local to global: A graph {RAG} approach to query-focused summarization.
\newblock \emph{arXiv preprint arXiv:2404.16130}, 2024.

\bibitem{hsieh2024ruler}
Hsieh, C.-P., Sun, S., Kriman, S., Acharya, S., Rekesh, D., Jia, F., Zhang, Y., and Ginsburg, B.
\newblock {RULER}: What's the real context size of your long-context language models?
\newblock In \emph{Conference on Language Modeling (COLM)}, 2024.

\bibitem{hu2025memoryagentbench}
Hu, Y., Wang, Y., and McAuley, J.
\newblock Evaluating memory in {LLM} agents via incremental multi-turn interactions.
\newblock In \emph{The Fourteenth International Conference on Learning Representations (ICLR)}, 2026.

\bibitem{izacard2022contriever}
Izacard, G., Caron, M., Hosseini, L., Riedel, S., Bojanowski, P., Joulin, A., and Grave, E.
\newblock Unsupervised dense information retrieval with contrastive learning.
\newblock \emph{Transactions on Machine Learning Research}, 2022.

\bibitem{karpathy2025wiki}
Karpathy, A.
\newblock {LLM} wiki.\newline
\url{https://gist.github.com/karpathy/442a6bf555914893e9891c11519de94f}, 2026.

\bibitem{khattab2024dspy}
Khattab, O., Singhvi, A., Maheshwari, P., Zhang, Z., Santhanam, K., Vardhamanan A, S., Haq, S., Sharma, A., Joshi, T., Moazam, H., Miller, H., Zaharia, M., and Potts, C.
\newblock {DSPy}: Compiling declarative language model calls into state-of-the-art pipelines.
\newblock In \emph{The Twelfth International Conference on Learning Representations (ICLR)}, 2024.

\bibitem{lu2024bm25s}
L\`u, X.~H.
\newblock {BM25S}: Orders of magnitude faster lexical search via eager sparse scoring.
\newblock \emph{arXiv preprint arXiv:2407.03618}, 2024.

\bibitem{maharana2024locomo}
Maharana, A., Lee, D.-H., Tulyakov, S., Bansal, M., Barbieri, F., and Fang, Y.
\newblock Evaluating very long-term conversational memory of {LLM} agents.
\newblock In \emph{Proceedings of the 62nd Annual Meeting of the Association for Computational Linguistics (ACL)}, pp.\ 13851--13870, 2024.

\bibitem{modarressi2025nolima}
Modarressi, A., Deilamsalehy, H., Dernoncourt, F., Bui, T., Rossi, R.~A., Yoon, S., and Sch{\"u}tze, H.
\newblock {NoLiMa}: Long-context evaluation beyond literal matching.
\newblock In \emph{Proceedings of the 42nd International Conference on Machine Learning (ICML)}, 2025.

\bibitem{packer2023memgpt}
Packer, C., Wooders, S., Lin, K., Fang, V., Patil, S.~G., Stoica, I., and Gonzalez, J.~E.
\newblock {MemGPT}: Towards {LLM}s as operating systems.
\newblock \emph{arXiv preprint arXiv:2310.08560}, 2023.

\bibitem{rasmussen2025zep}
Rasmussen, P., Paliychuk, P., Beauvais, T., Ryan, J., and Chalef, D.
\newblock Zep: A temporal knowledge graph architecture for agent memory.
\newblock \emph{arXiv preprint arXiv:2501.13956}, 2025.

\bibitem{ryokoai2023sharegpt}
RyokoAI.
\newblock {ShareGPT52K}.
\newblock \url{https://huggingface.co/datasets/RyokoAI/ShareGPT52K}, 2023.

\bibitem{tan2025membench}
Tan, H., Zhang, Z., Ma, C., Chen, X., Dai, Q., and Dong, Z.
\newblock {MemBench}: Towards more comprehensive evaluation on the memory of {LLM}-based agents.
\newblock In \emph{Findings of the Association for Computational Linguistics: ACL 2025}, pp.\ 19336--19352, 2025.

\bibitem{wu2024longmemeval}
Wu, D., Wang, H., Yu, W., Zhang, Y., Chang, K.-W., and Yu, D.
\newblock {LongMemEval}: Benchmarking chat assistants on long-term interactive memory.
\newblock In \emph{The Thirteenth International Conference on Learning Representations (ICLR)}, 2025.

\bibitem{zhang2024survey}
Zhang, Z., Dai, Q., Bo, X., Ma, C., Li, R., Chen, X., Zhu, J., Dong, Z., and Wen, J.-R.
\newblock A survey on the memory mechanism of large language model-based agents.
\newblock \emph{ACM Transactions on Information Systems}, 43(6):1--47, 2025.

\bibitem{yang2018hotpotqa}
Yang, Z., Qi, P., Zhang, S., Bengio, Y., Cohen, W.~W., Salakhutdinov, R., and Manning, C.~D.
\newblock {HotpotQA}: A dataset for diverse, explainable multi-hop question answering.
\newblock In \emph{Proceedings of the Conference on Empirical Methods in Natural Language Processing (EMNLP)}, 2018.

\bibitem{zhong2023mquake}
Zhong, Z., Wu, Z., Manning, C.~D., Potts, C., and Chen, D.
\newblock {MQuAKE}: Assessing knowledge editing in language models via multi-hop questions.
\newblock In \emph{Proceedings of the Conference on Empirical Methods in Natural Language Processing (EMNLP)}, 2023.

\bibitem{cohen2024ripple}
Cohen, R., Biran, E., Yoran, O., Globerson, A., and Geva, M.
\newblock Evaluating the ripple effects of knowledge editing in language models.
\newblock \emph{Transactions of the Association for Computational Linguistics}, 12:283--298, 2024.

\bibitem{trivedi2022musique}
Trivedi, H., Balasubramanian, N., Khot, T., and Sabharwal, A.
\newblock {MuSiQue}: Multihop questions via single hop question composition.
\newblock \emph{Transactions of the Association for Computational Linguistics}, 10:539--554, 2022.

\bibitem{lewis2020rag}
Lewis, P., Perez, E., Piktus, A., Petroni, F., Karpukhin, V., Goyal, N., Küttler, H., Lewis, M., Yih, W., Rocktäschel, T., Riedel, S., and Kiela, D.
\newblock Retrieval-augmented generation for knowledge-intensive {NLP} tasks.
\newblock In \emph{Advances in Neural Information Processing Systems (NeurIPS)}, 2020.

\bibitem{zheng2023llmjudge}
Zheng, L., Chiang, W.-L., Sheng, Y., Zhuang, S., Wu, Z., Zhuang, Y., Lin, Z., Li, Z., Li, D., Xing, E.~P., Zhang, H., Gonzalez, J.~E., and Stoica, I.
\newblock Judging {LLM}-as-a-judge with {MT-Bench} and {Chatbot Arena}.
\newblock In \emph{Advances in Neural Information Processing Systems (NeurIPS)}, 2023.

\end{thebibliography}
}
\end{document}